\documentclass[runningheads]{llncs}
\usepackage{graphicx}
\usepackage{subcaption}
\begin{document}
\title{Intensity and Texture Correction of Omnidirectional Image Using Camera Images\\ for Indirect Augmented Reality}
%
%
\author{Hakim Ikebayashi\and
Norihiko Kawai\orcidID{0000-0002-7859-8407}}
\authorrunning{H. Ikebayashi et al.}
%
\institute{Graduate School of Information Science and Technology, \\Osaka Institute of Technology 
\\1-79-1 Kitayama, Hirakata, Osaka 573-0196, Japan\\
\email{norihiko.kawai@oit.ac.jp}}
\maketitle              
\begin{abstract}
Augmented reality (AR) using camera images in mobile devices is becoming popular for tourism promotion. However, obstructions such as tourists appearing in the camera images may cause the camera pose estimation error, resulting in CG misalignment and reduced visibility of the contents. To avoid this problem, Indirect AR (IAR), which does not use real-time camera images, has been proposed. In this method, an omnidirectional image is captured and virtual objects are synthesized on the image in advance. Users can experience AR by viewing a scene extracted from the synthesized omnidirectional image according to the device's sensor. This enables robustness and high visibility. However, if the weather conditions and season in the pre-captured 360 images differs from the current weather conditions and season when AR is experienced, the realism of the AR experience is reduced. To overcome the problem, we propose a method for correcting the intensity and texture of a past omnidirectional image using camera images from mobile devices. We first perform semantic segmentation. We then reproduce the current sky pattern by panoramic image composition and inpainting. For the other areas, we correct the intensity by histogram matching. In experiments, we show the effectiveness of the proposed method using various scenes. 

\end{abstract}
\section{Introduction}
Augmented reality (AR) using camera images from mobile devices is becoming popular for the purpose of tourism promotion. However, tourists appearing in the image sequence captured by the camera can cause the camera pose estimation error, resulting in CG misalignment and reduced visibility of the contents. To avoid this problem, Indirect AR (IAR) \cite{IndirectAR}, which does not use real-time camera images, has been proposed. In this method, an omnidirectional image is captured with an omnidirectional camera, and virtual objects are synthesized on the image in advance. Users can experience AR by viewing a scene extracted from the synthesized omnidirectional image according to the device's sensor. This enables robustness and high visibility. However, if the weather conditions and season in the pre-captured omnidirectional image differ from the current weather conditions and season when AR is experienced, the realism of the AR experience is reduced.

To overcome the problem, we propose a method for correcting the intensity and texture of a past omnidirectional image using camera images acquired by a mobile device to match the current scene as a calibration before users experience IAR. We first perform semantic segmentation \cite{SemanticSegmentation}. We then reproduce the current sky pattern by panoramic image composition and inpainting \cite{inpainting}. For the other areas, we correct the intensity by histogram matching. The corrected omnidirectional image is displayed on the screen using the mobile device's sensor to reproduce the current realistic view in IAR.

\section{Related Work}
Many studies have been conducted in the tourism field with AR technology. In this section, we first describe systems for AR-based tourism using real-time camera images, and then discuss its problems. Next, we introduce application examples of Indirect AR, which compensates for the shortcomings of AR using camera images, and describe the issues of Indirect AR and the conventional approaches that have attempted to solve them.

\subsection{AR Experiences in Tourism}
AR is used for various purposes in tourism, especially for guiding people \cite{ARmuseum} and visually reconstructing buildings of the past in historical sites \cite{ARsightseeingXRSample2018}. For example, Amakawa \cite{ARsightseeingXRSample2017} developed The New Philadelphia AR Tour, a mobile application that allows the visitor to walk through historical building reconstruction using AR. An AR tour application that reconstructs information on underground structures with 3D models has also been developed \cite{ARtour}.

In such applications, camera pose estimation is significant to place virtual objects in the proper 3D positions. The camera pose estimation has been performed by solving PnP problems using AR markers \cite{Kato1999} and natural feature points \cite{Taketomi2011}, or by fusing Visual-SLAM \cite{SLAM2022} and sensors such as GNSS and gyroscope. However, in crowded tourist areas, people often appear in the camera image, causing the camera pose estimation to fail. In addition, the objects that a user wants to capture does not appear in camera images due to occlusion, which spoils the AR experience. For this problem, although diminished reality, which removes unwanted objects in the camera images, has also been studied \cite{Kawai2016,Kawai2017,transformr2021,DRsurvey}, it is still difficult for existing methods to naturally remove many people in camera images.

\subsection{Example Usage of Indirect AR and Issues}

IAR, which solves the above problems of AR using real-time camera images, has been proposed \cite{IndirectAR}. Since IAR presents a part of the omnidirectional image, in which unwanted objects do not appear and virtual objects are aligned, according to the user's device pose, the user can enjoy AR without being affected by surrounding tourists at all. Because of its robustness and visibility, there are several examples of its use for tourism.
For example, Gimeno \cite{IARindoor} introduces a system that combines AR and IAR that can work on mobile devices held by visitors to improve the museum visitor experience.
Suganuma \cite{IAR2023} developed an IAR system that allows visitors to experience electronically reconstructed ancient buildings at historical sites.

However, IAR that uses a past omnidirectional image has two major problems that are different from those of AR using real-time camera images. One is that the viewpoint of the AR image cannot be changed even when the use moves the mobile device. The other is that the actual landscape and the one presented on the mobile device differ when the current season and weather conditions differ from those when the omnidirectional image was captured in the past. These problems degrade the quality of the AR experience.

For the former problem, for example, the method in \cite{IARmotionparallax} estimates camera motion by optical flow so that the method can work even when there are many people around, and then generates the novel viewpoint image by viewpoint-dependent texture mapping to represent motion parallax. In addition, for free viewpoint image generation, many NeRF-based methods \cite{NeRF,kerbl3Dgaussians} have been studied in recent years, among which a method that can generate free viewpoint images for any direction has been proposed \cite{VRNeRF}, which can be applied to IAR.

For the latter problem, Okura et al. \cite{IAR2017} proposed a method in which omnidirectional images are captured at various times in advance and stored in a database, and when experiencing IAR, one of the images that is closest in appearance to the image captured with a mobile device by the user is selected. This allows users to experience IAR using the image with the lighting condition close to the current one. However, the cost of building a database by acquiring omnidirectional images at various times is quite high. On the other hand, methods that use deep learning to translate the season, weather, and day/night of the landscape in an image have been studied using GAN (Generative Adversarial Network) \cite{CycleGAN2017,WeatherGAN,Son2023}, and it may be possible to apply these methods to IAR. However, they are highly dependent on training data and do not always exactly reproduce the current landscape.

This study focuses on the latter optical inconsistency problem. Unlike the conventional methods, the proposed method has no database construction cost because it uses a single omnidirectional image. In addition, although the proposed method translates a past omnidirectional image similarly to the GAN-based methods, it uses camera images from the user's mobile device to translate the omnidirectional image so that it is closer to the current landscape.

\section{Proposed Method}
We first describe an overview of conventional IAR \cite{IndirectAR}, and an overview of the proposed method. We then describe the details of the proposed method.

\subsection{Overview of Indirect AR}
IAR can be divided into a pre-processing phase and an AR experience phase. As a pre-processing phase, an omnidirectional image is captured in advance as a preparation at the point where we want users to experience AR. The pre-captured image is then mapped onto a sphere in a virtual space, and a virtual camera is placed at the center of the sphere. Next, virtual objects are placed in the sphere so that the positional relationship between the real world and virtual objects are consistent in the field of view from the center of the sphere. By rendering the omnidirectional image with the center of the sphere as the viewpoint, the virtual objects are composited into the omnidirectional image. Or, the sphere as the real world and the virtual objects inside the sphere are left as they are and used in the system described below.

As an AR experience phase, the virtual camera at the center of the sphere is rotated according to the pose obtained from the mobile device's sensors such as electronic compass, gyro sensor, and acceleration sensor. The image captured by the virtual camera is presented to the user on the screen of the mobile device.

\subsection{Overview of Proposed Method}
In the proposed method, the texture and intensity in the omnidirectional image captured in advance is corrected using a panoramic image taken and generated by the user so that a scene in the image becomes close to the present one as a calibration before the actual AR experience at the AR experience phase.

\begin{figure}[bt]
\begin{center}
\includegraphics[width=0.8\textwidth]{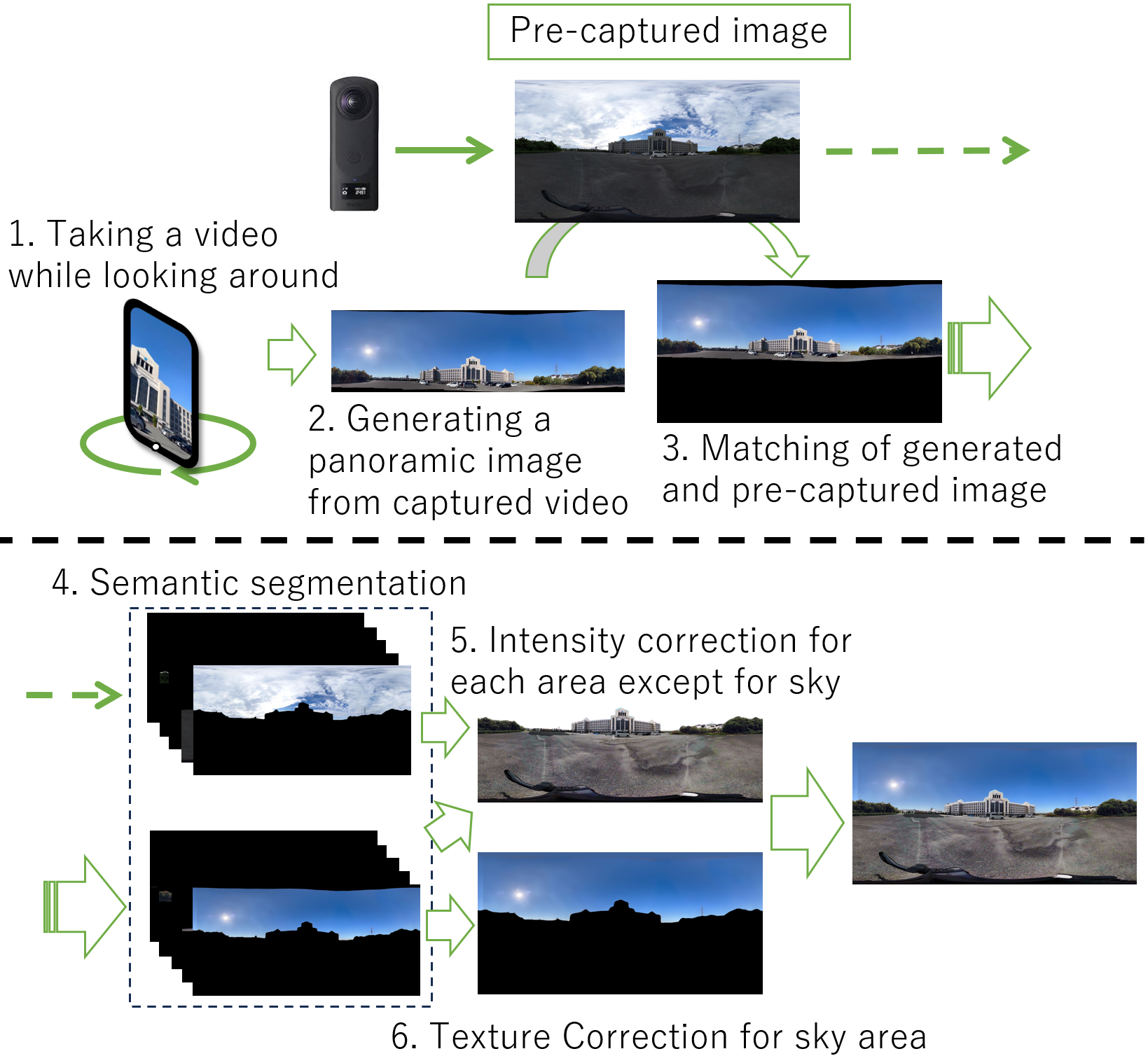}
\caption{Flow of the proposed method.}
\label{fig:flow}
\end{center}
\end{figure}

Figure \ref{fig:flow} shows the flow of the proposed method conducted before a user experiences AR at the AR experience phase. (1) The user first takes a video while looking around at the fixed point using the camera of the mobile device used for the AR. (2) A panoramic image is then generated from the captured video. (3) Feature point extraction and matching using AKAZE \cite{AKAZE} are performed on the pre-captured omnidirectional image and the generated panoramic image. The generated panoramic image is then transformed using the scaling and translation matrix calculated from the correspondence of feature points by matching, to obtain the same appearance as the pre-captured image. (4) Semantic segmentation \cite{SemanticSegmentation} is performed on the pre-captured image and the generated panoramic image. (5) For each of the same category of the pre-captured image and the generated panoramic image, except for the sky area, the color tone of the pre-captured image is converted to the color tone of the generated panoramic image using histogram matching. (6) For the sky area, the generated panoramic image is basically copied to the pre-captured image. In addition, the areas not covered by the panorama are completed by inpainting \cite{inpainting}. (7) Finally, the omnidirectional image with the sky texture and the intensity in the other areas corrected is used for IAR. In the following, we describe the intensity and texture correction of pre-captured omnidirectional images using the proposed method in detail.

\subsection{Taking a Video while Looking Around}
As step (1), a user takes a video while looking around at the fixed point using the camera of the mobile device used for the AR experience. Specifically, the mobile device's camera is set to a wide angle of view and the user rotates the camera 360 degrees so that as many objects and large sky areas as possible can be captured. Figure \ref{fig:image45fps} shows example frames extracted from a video.

\subsection{Generating a Panoramic Image from Captured Video}

\begin{figure}[tb]
\begin{center}
\begin{minipage}{0.13\textwidth}
\includegraphics[width=\textwidth]{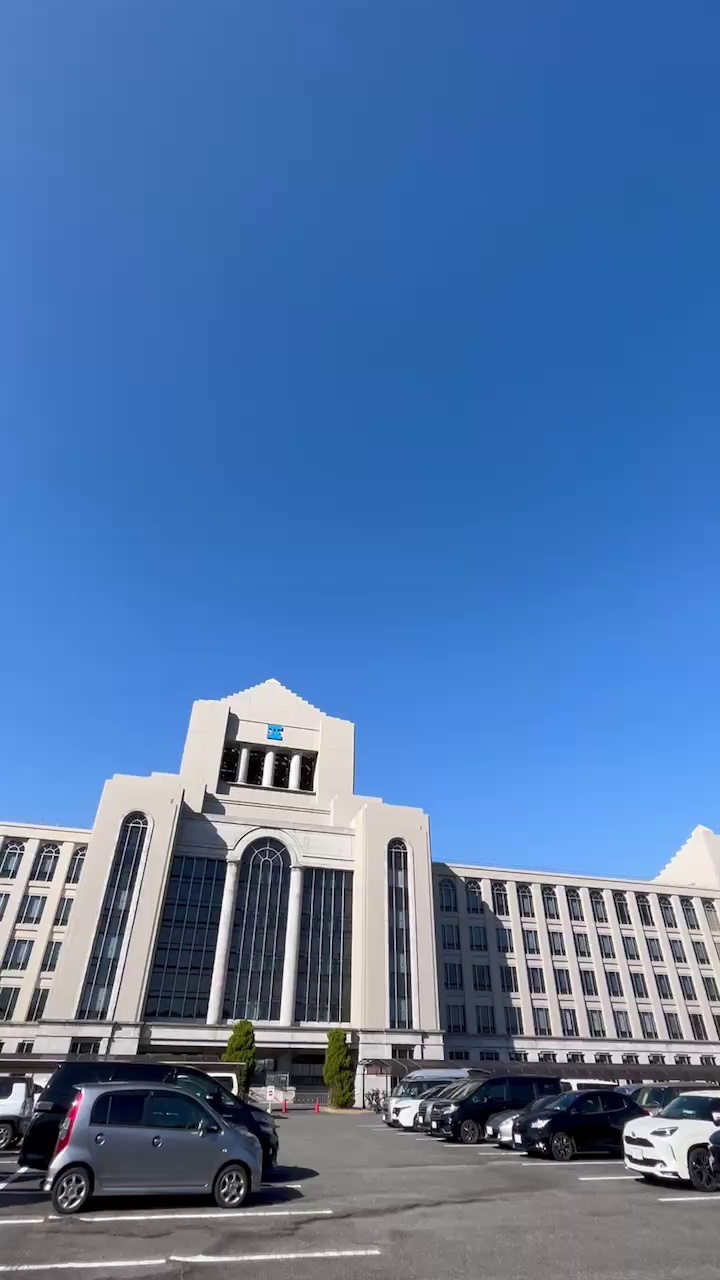}
\end{minipage}
\begin{minipage}{0.13\textwidth}
\includegraphics[width=\textwidth]{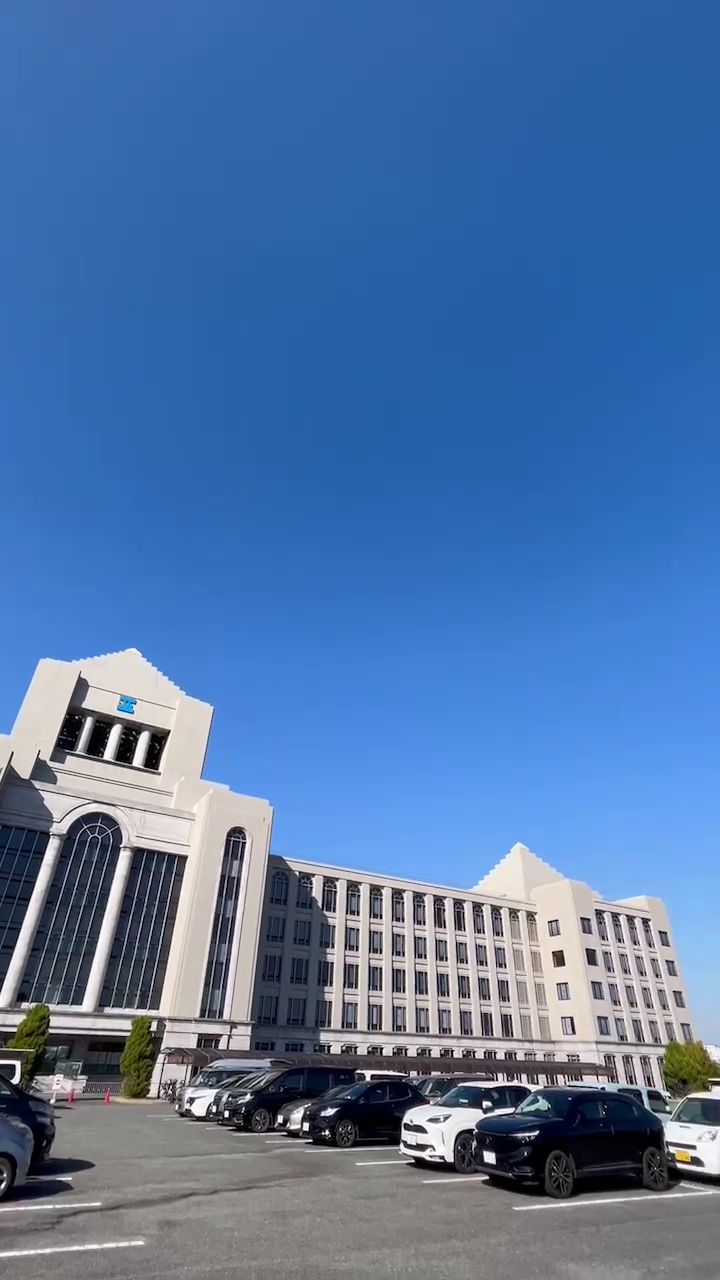}
\end{minipage}
\caption{Example images extracted from video.}
\label{fig:image45fps}
\end{center}
\end{figure}

\begin{figure}[tb]
\begin{center}
\includegraphics[width=0.6\textwidth]{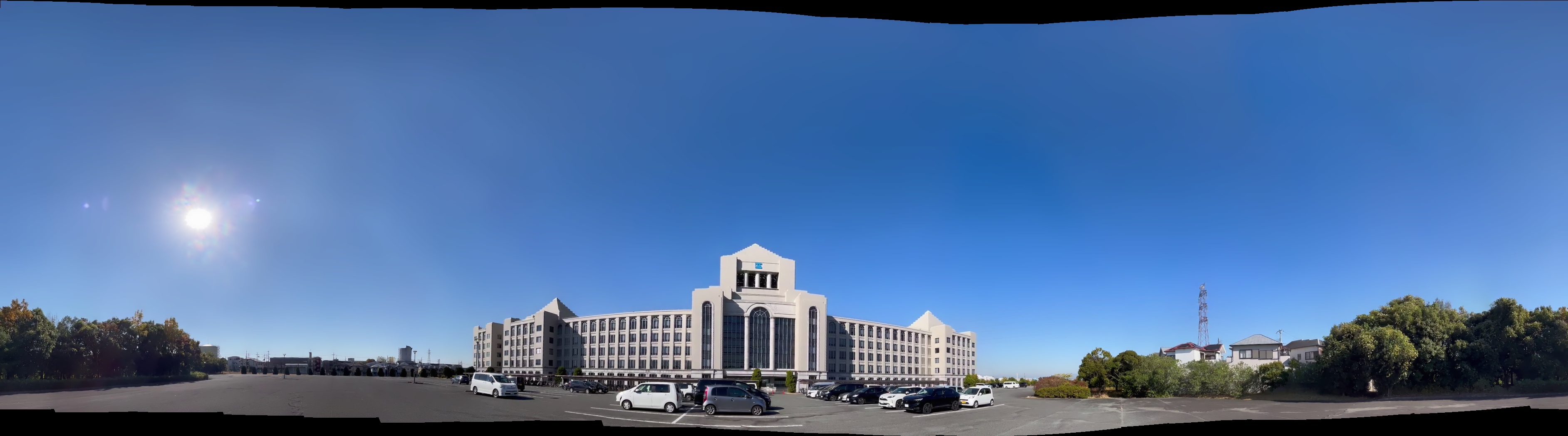}
\caption{Panoramic image generated with equirectangular projection.}
\label{fig:image2panorama}
\end{center}
\end{figure}

As step (2), a panoramic image is generated from the video. Here, feature points are detected in the images extracted from the video, and the images are stitched together by mapping the images onto a sphere with feature matching.
Figure \ref{fig:image2panorama} shows a panoramic image generated from a video including Fig. \ref{fig:image45fps}.
The panoramic image is generated as an equirectangular projection image.

\subsection{Matching of Generated and Pre-captured Images}

\begin{figure}[bt]
\begin{center}
\includegraphics[width=0.48\textwidth]{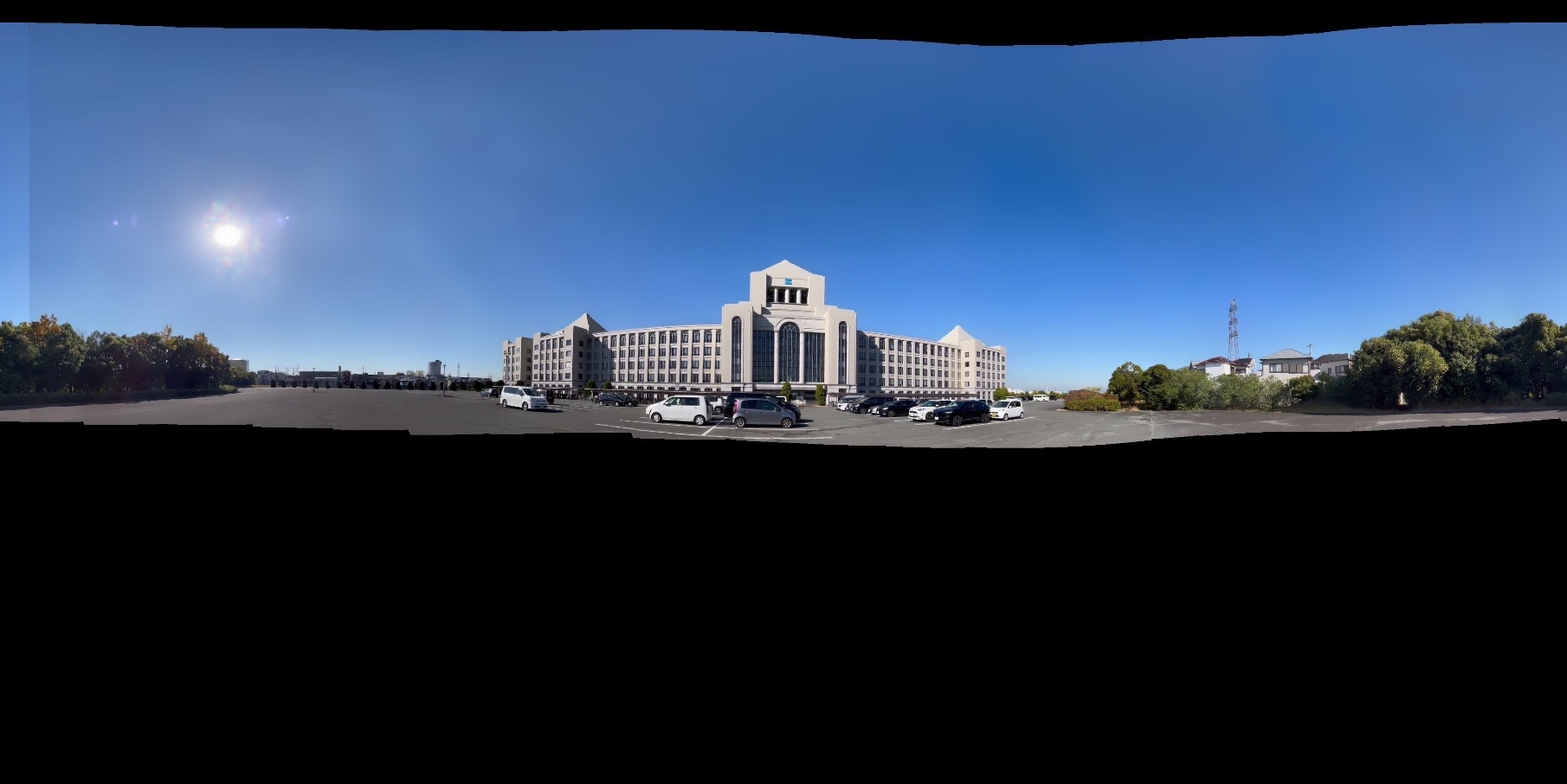}
\caption{Transformed panoramic image.}
\label{fig:transformed}
\end{center}
\end{figure}

\begin{figure}[bt]
\begin{center}
\includegraphics[width=0.48\textwidth]{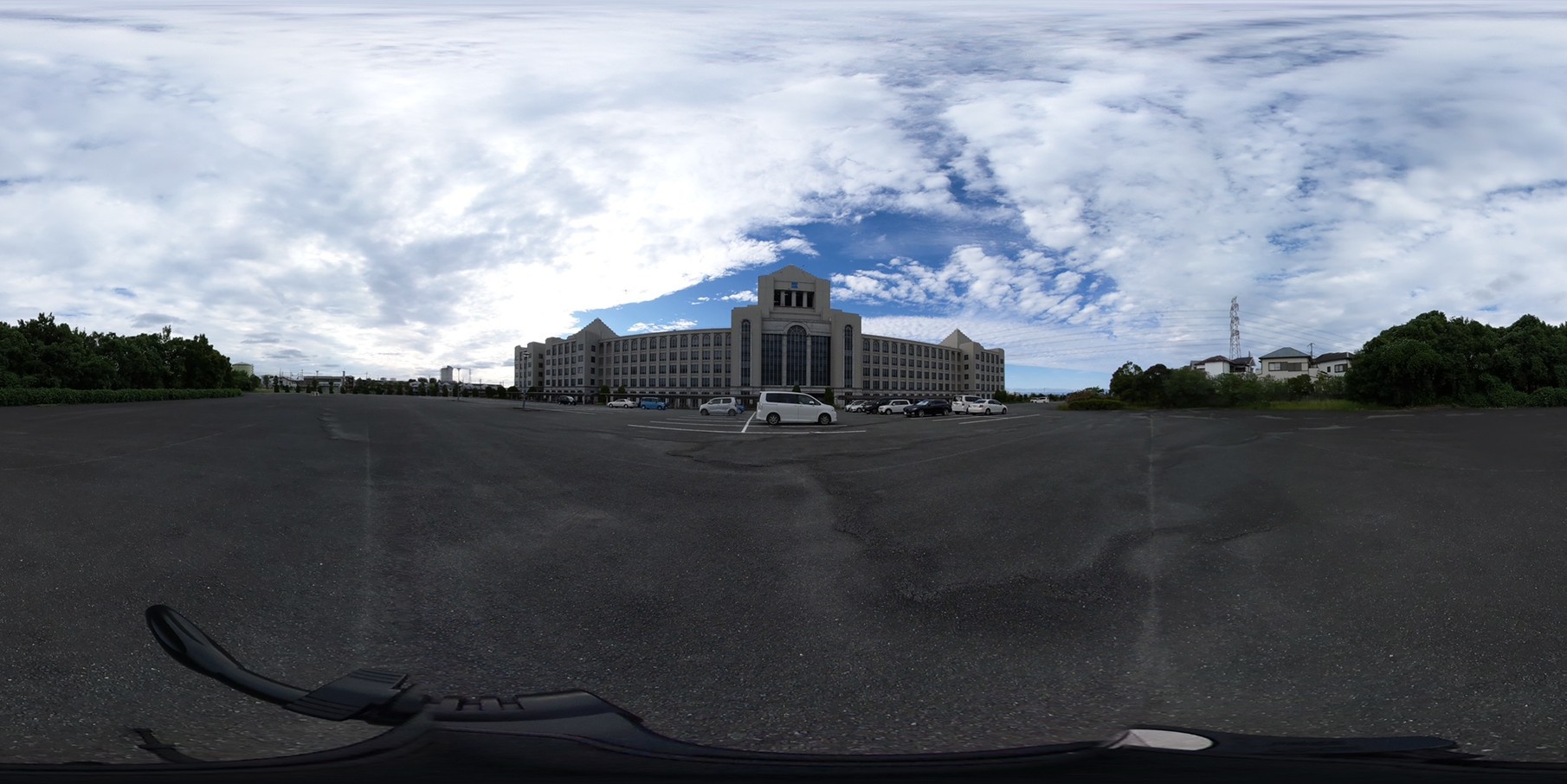}
\caption{Pre-captured omnidirectional image.}
\label{fig:precaptured}
\end{center}
\end{figure}

As step (3), the generated panorama image is transformed so that the appearance is the same as the pre-captured omnidirectional image. For this, AKAZE \cite{AKAZE} is first used to extract and match feature points on the pre-capture image and the generated image. The generated image is then transformed using the scaling and translation matrix calculated from the correspondence of feature points by matching.
This transformation matrix is calculated by excluding outliers in the pairs of the detected feature points using RANSAC \cite{ransac}, and then using the least-squares method for the inlier correspondences.
In addition, taking into account that the scene being photographed is omnidirectional, the part of the image that extends beyond on the edges of the image are made to appear from the opposite side of the image during the transformation. Here, when a panoramic image is wider than 360 degrees, the edge areas of the panoramic image may overlap after the transformation. In such a case, the pixel values in one of the areas are overwritten. In addition, if the panoramic and the pre-captured images have a large misalignment, a single conversion may not work properly. Therefore, the above process is repeated several times.
Figure \ref{fig:transformed} shows an example of the image transformed from Fig. \ref{fig:image2panorama} so that the appearance becomes the same as Fig. \ref{fig:precaptured}, which is an example of the pre-captured omnidirectional image. 

\subsection{Semantic Segmentation}
\begin{figure}[tb]
\begin{center}
\begin{minipage}{0.48\textwidth}
\includegraphics[width=\textwidth]{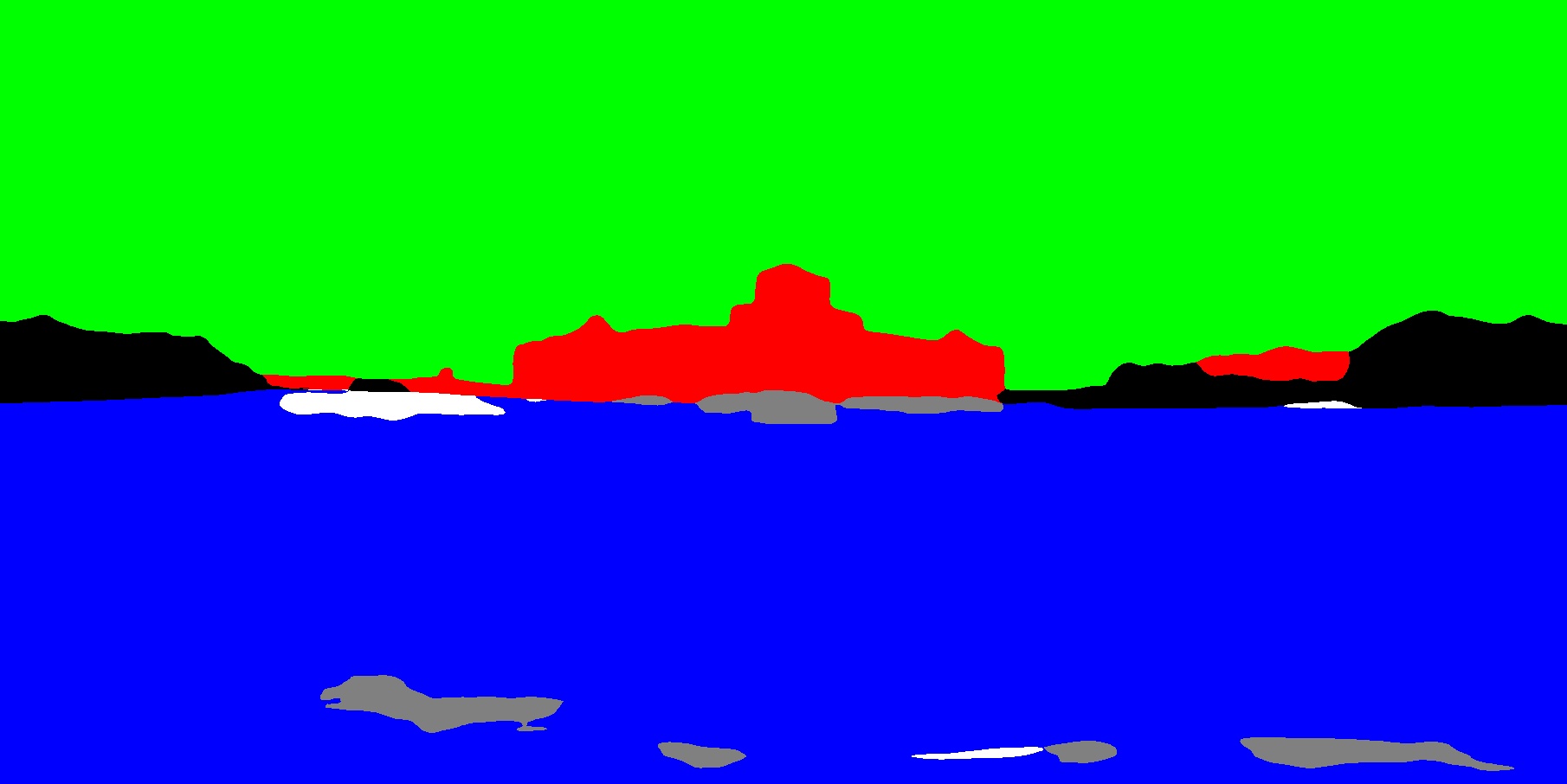}
\subcaption{Labeling of pre-captured image}
\end{minipage}
\begin{minipage}{0.48\textwidth}
\includegraphics[width=\textwidth]{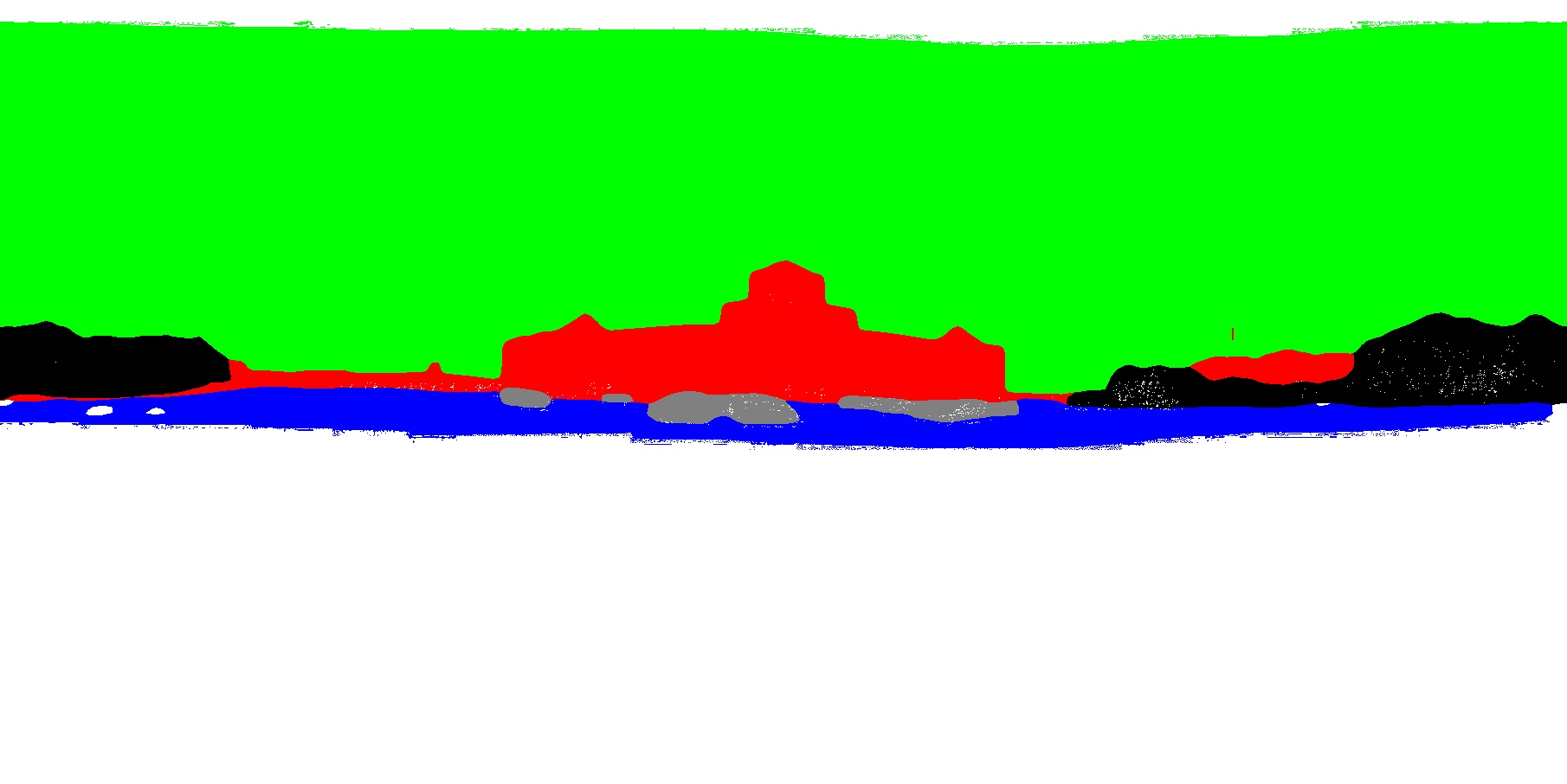}
\subcaption{Labeling of generated image}
\end{minipage}
\caption{Lableling images.}
\label{fig:labelingimage}
\end{center}
\end{figure}

For the intensity and texture correction of the pre-captured image to be done in the later process, the pre-captured and generated images are divided into five categories of areas including sky area using semantic segmentation \cite{SemanticSegmentation} as shown in Fig. \ref{fig:labelingimage}, as step (4). 
A scene may be able to be segmented into areas of more categories. 
However, there is a possibility that some categories appear only in one of the pre-captured and generated images, and in this case, the correction may fail. 
For this reason, this method extracts only the sky and the top four categories included in the generated image.
In the following, we describe methods for correcting intensity and texture in non-sky and sky areas separately.

\subsection{Intensity Correction for Each Area except for Sky}

\begin{figure}[tb]
\begin{center}
\includegraphics[width=1.0\textwidth]{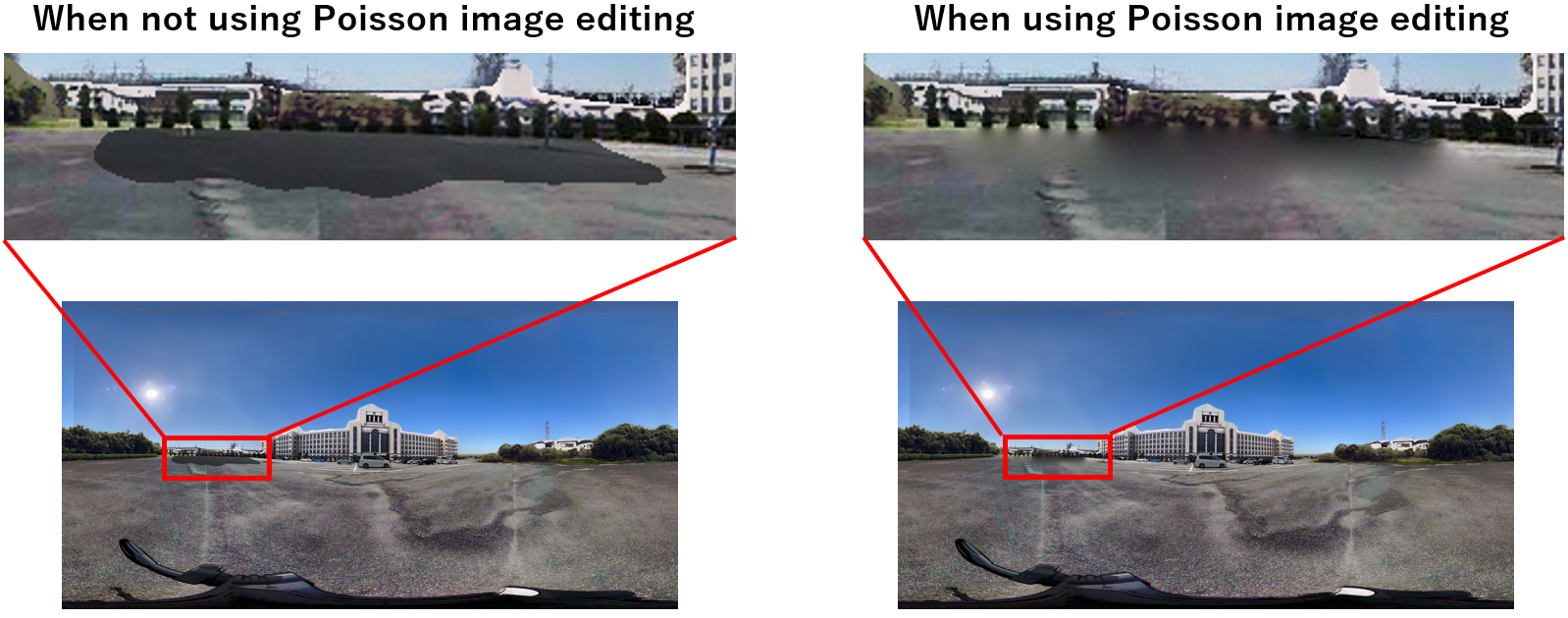}
\caption{Before and after Poisson Image Editing.}
\label{fig:poisson}
\end{center}
\end{figure}

As step (5), the intensity of the pre-captured image is corrected using the generated image by histogram matching.
Specifically, normalized cumulative histograms are first created for each category in both the pre-captured and transformed images. 
A transformation function of the pixel values is obtained for each category and each RGB separately so that the normalized cumulative histogram of the pre-captured image matches that of the transformed image. 
The intensity of the pre-captured image is corrected using the transformation function.
Finally, the intensity of the ramaining unclassified areas are adjusted using Poisson Image Editing \cite{Perez2003} to match the surrounding intensity, as shown in the Fig. \ref{fig:poisson}.
This process reduces the difference in luminance between the areas in the top five categories and the remaining areas.

\begin{figure}[bt]
\begin{center}
\includegraphics[width=0.48\textwidth]{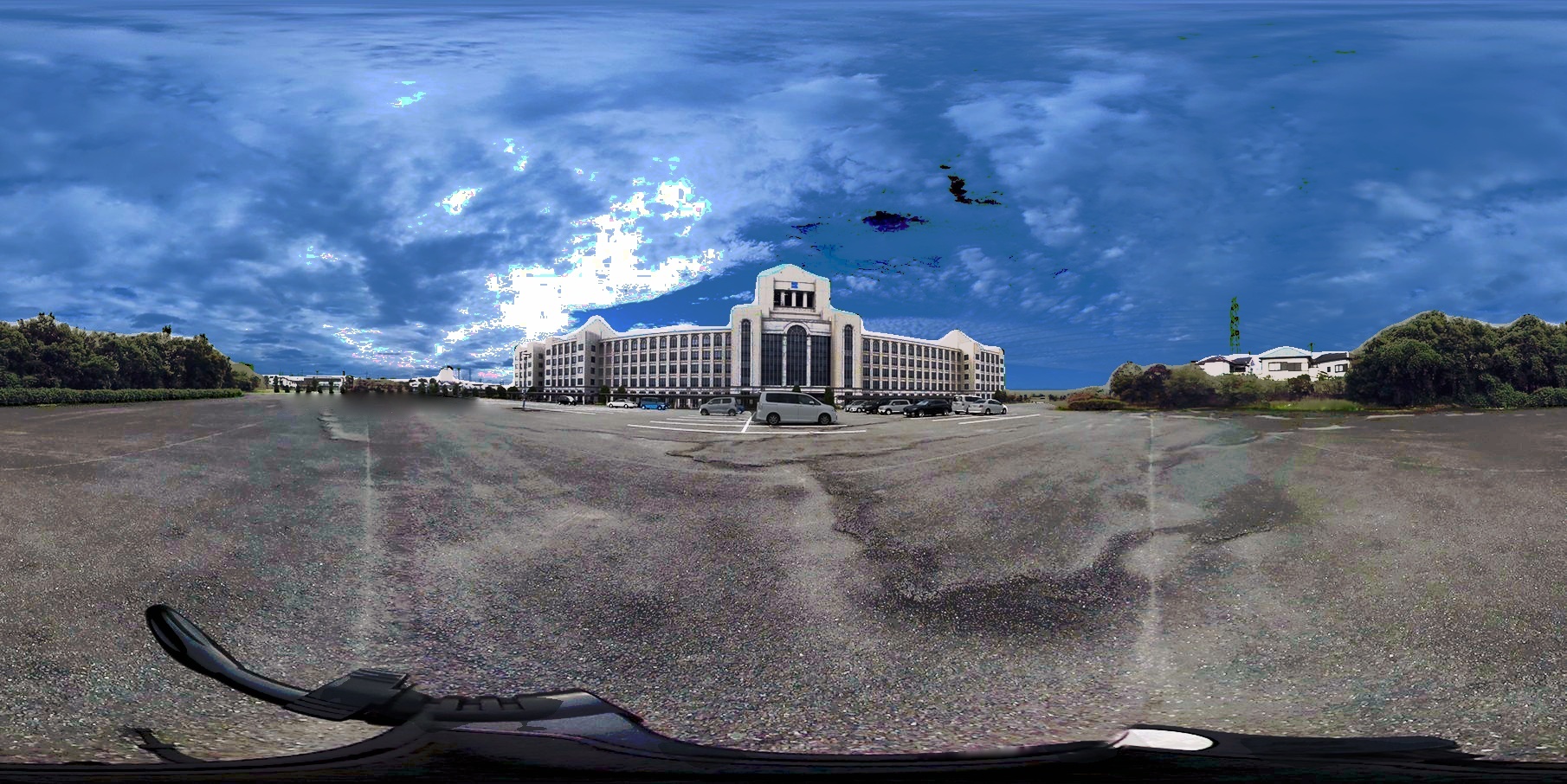}
\caption{Pre-captured image after intensity correction.}
\label{fig:histgrammatching}
\end{center}
\end{figure}

Figure \ref{fig:histgrammatching} shows the result of intensity correction so that the normalized cumulative histograms obtained from Fig. \ref{fig:precaptured} matches those from Fig.\ref{fig:transformed}.
As we can see from the comparison with Figs. \ref{fig:histgrammatching} and \ref{fig:transformed}, though the intensity in areas except for the sky becomes much closer, the texture of the sky area is still quite different. Therefore, a different approach is used for the sky area.

\subsection{Texture Correction for Sky Area}
As step (6), for the sky area, the generated and transformed panoramic image is basically copied to the pre-captured omnidirectional image. However, the panoramic image does not necessarily cover the entire sky area. Therefore, inpainting \cite{inpainting} is also used to fill in the missing areas. 

\begin{figure}[tb]
\begin{center}
\begin{minipage}{0.27\textwidth}
\includegraphics[width=\textwidth]{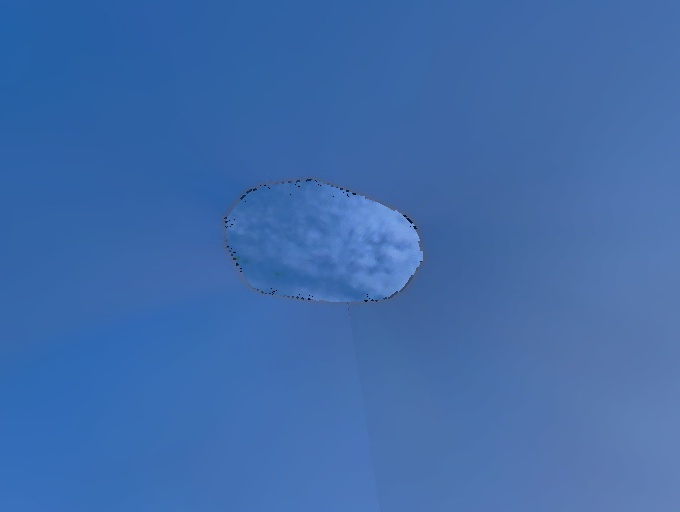}
\subcaption{Example of image before inpainting}
\end{minipage}
\begin{minipage}{0.27\textwidth}
\includegraphics[width=\textwidth]{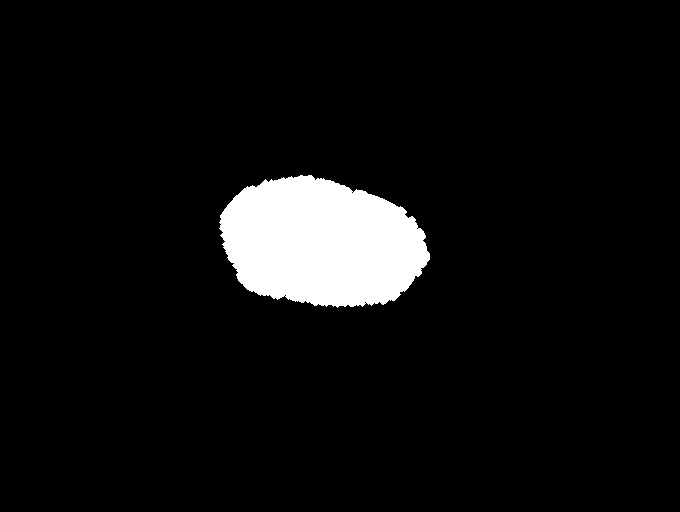}
\subcaption{Example of mask image of sky area}
\end{minipage}
\begin{minipage}{0.27\textwidth}
\includegraphics[width=\textwidth]{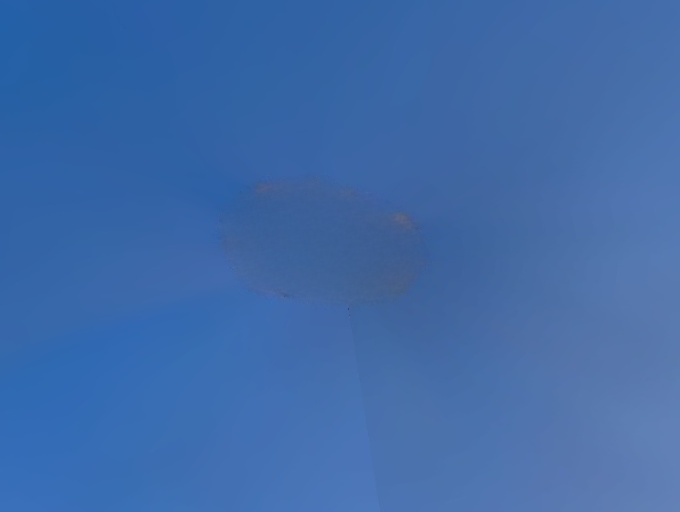}
\subcaption{Example of image after inpainting}
\end{minipage}
\caption{Cut out to perspective projection images.}
\label{fig:panorama2image}
\end{center}
\end{figure}

\begin{figure}[tb]
\begin{center}
\includegraphics[width=0.48\textwidth]{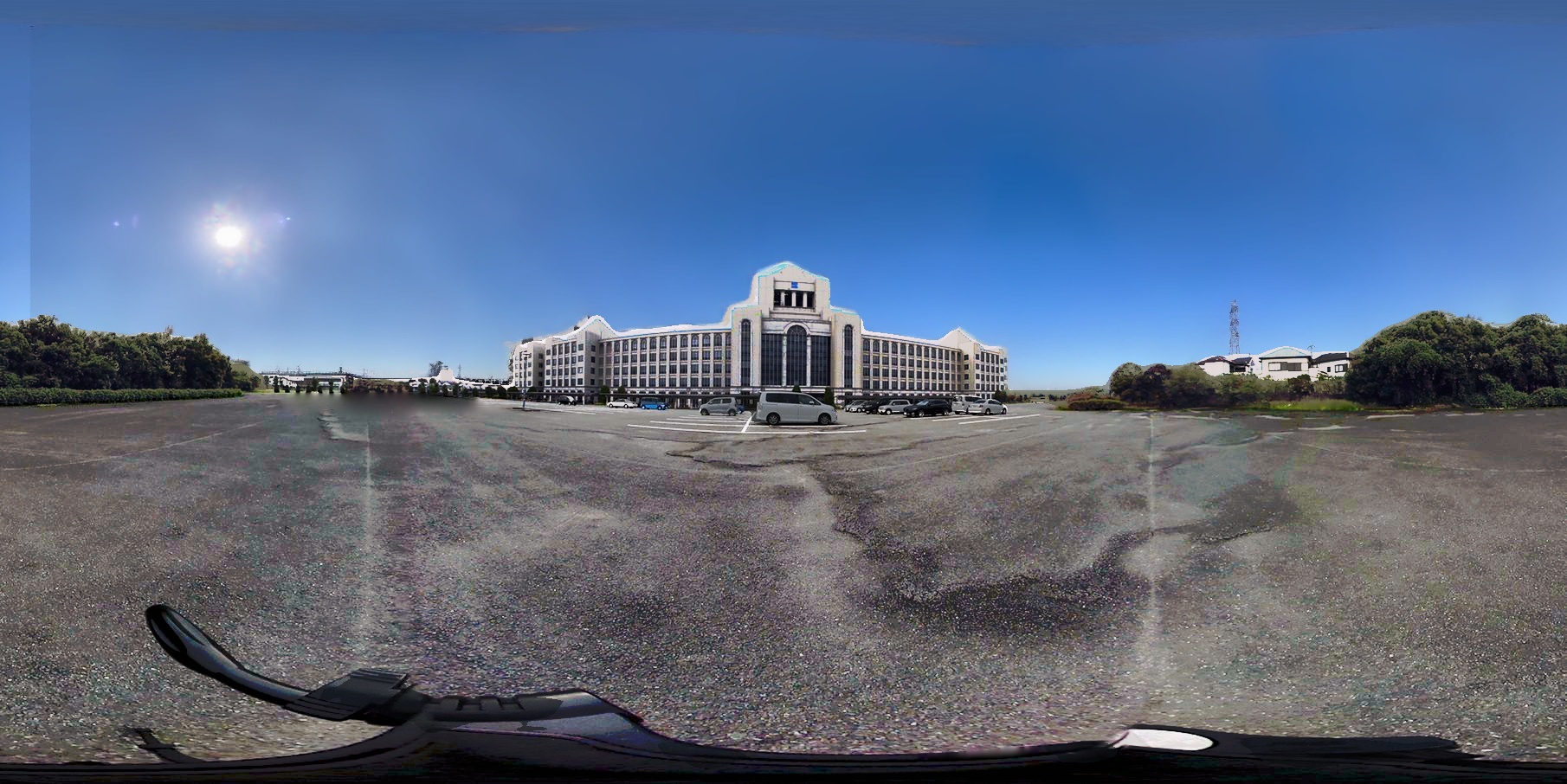}
\caption{Inpainting result of omnidirectional image.}
\label{fig:afterinpainting}
\end{center}
\end{figure}

Here, according to the generated image as shown in Fig. \ref{fig:transformed}, a mask image representing the missing area in the sky is created and the missing region is inpainted using the mask image. However, inpainting, such as the method \cite{inpainting}, is compatible with perspective projection images, but not with panoramic images with equirectangular projection in which the upper part is highly distorted. Therefore, the upper part of the panoramic image is converted to a perspective projection image, and inpainting is performed on that image. Figures \ref{fig:panorama2image}(a)(b) are the perspective color and mask images converted from the panoramic image as shown in Fig. \ref{fig:histgrammatching}, and the region in the color image corresponding to white in the mask image is inpainted as shown in Fig. \ref{fig:panorama2image}(c). The inpainted result is then converted back to the omnidirectional image as shown in Fig. \ref{fig:afterinpainting}.

\section{Experiments}
\begin{figure}[bt]
\begin{center}
\begin{minipage}{0.48\textwidth}
\includegraphics[width=\textwidth]{pic/pre-captured.jpg}
\subcaption{Location 1}
\end{minipage} \hspace{1mm}
\begin{minipage}{0.48\textwidth}
\includegraphics[width=\textwidth]{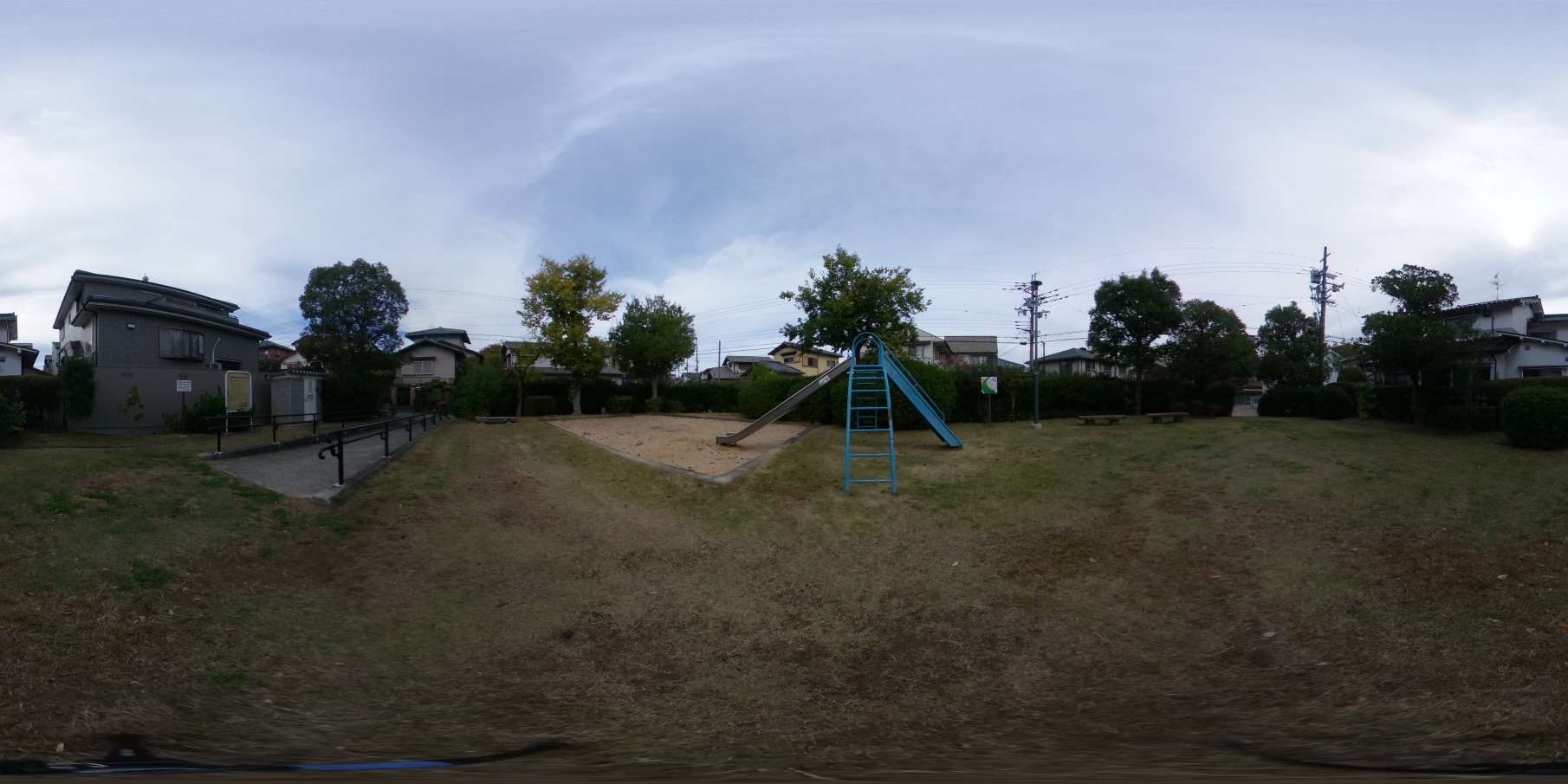}
\subcaption{Location 2}
\end{minipage}\\
\begin{minipage}{0.48\textwidth}
\includegraphics[width=\textwidth]{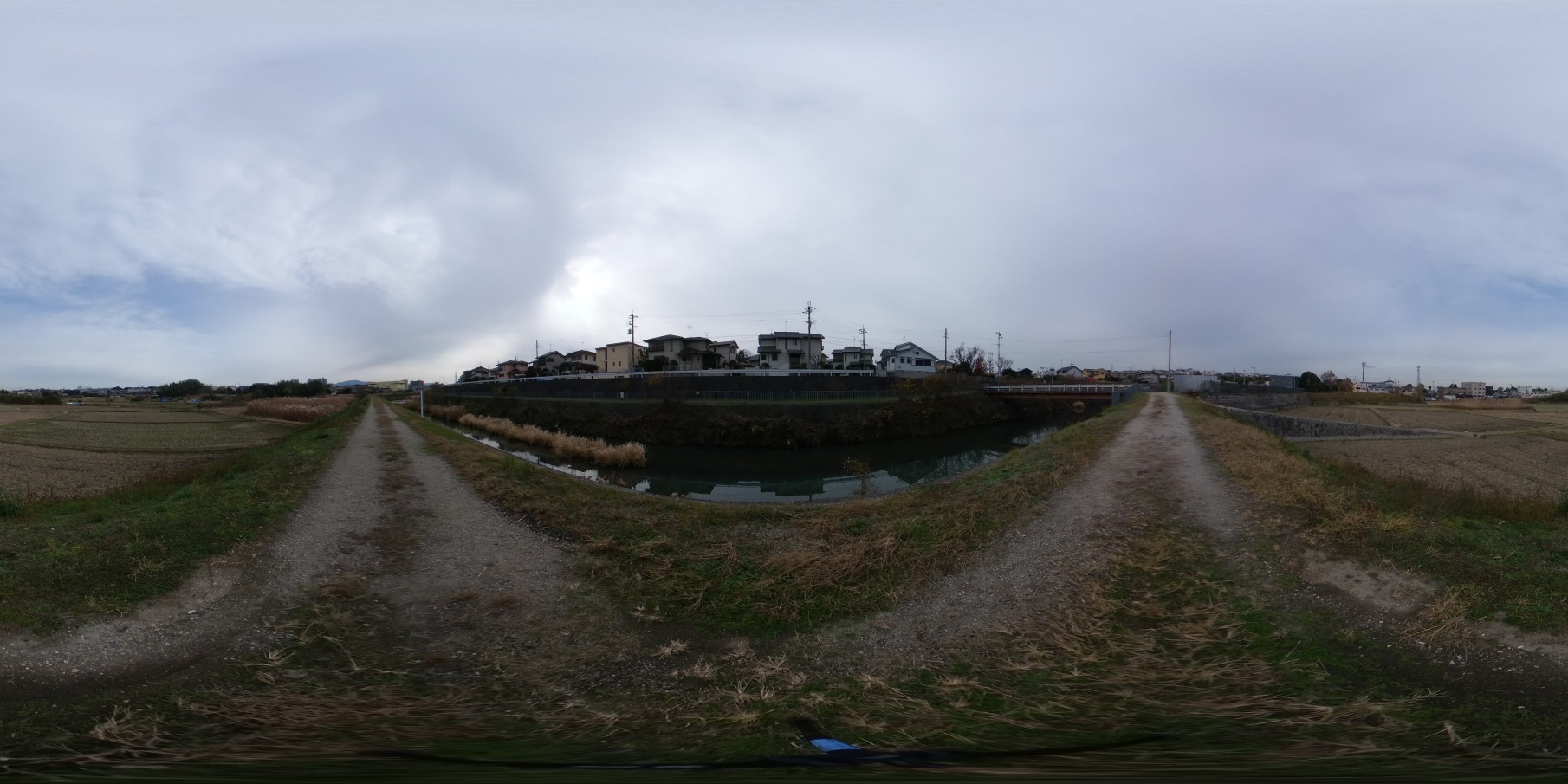}
\subcaption{Location 3}
\end{minipage}\\
\caption{Pre-captured omnidirectional images.}
\label{fig:pre-captureimages}
\end{center}
\end{figure}

To demonstrate the effectiveness of the proposed method, we conducted three experiments at three different locations, as shown in Fig. \ref{fig:pre-captureimages}, under different weather conditions and at different times of the day. We used RICOH THETA Z1 as an omnidirectional camera for pre-captured images, and iPhone 14 as a mobile device for capturing a video for panoramic images.
The resolution of the pre-captured images is 1810 $\times$ 906, and we took a video with 1080 $\times$ 1920 resolution for panoramic images.
We describe the experiments in the following.

\subsection{Experiment 1}
\begin{figure}[tp]
\begin{center}
\begin{minipage}{0.48\textwidth}
\includegraphics[width=\textwidth]{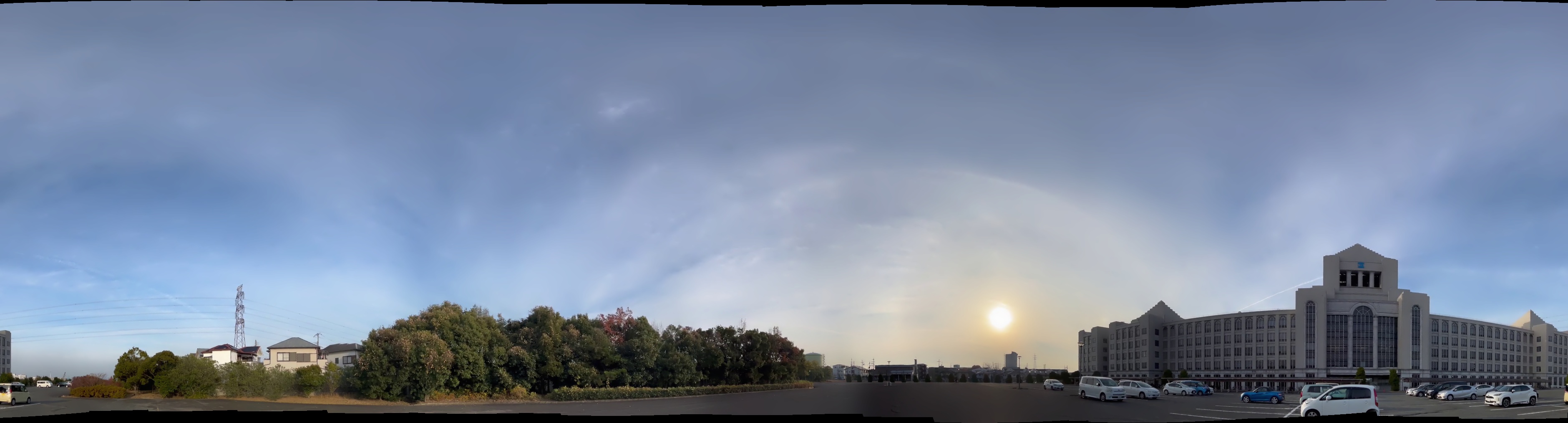}
\subcaption{Panoramic image in Sunny}
\end{minipage} \\
\begin{minipage}{0.48\textwidth}
\includegraphics[width=\textwidth]{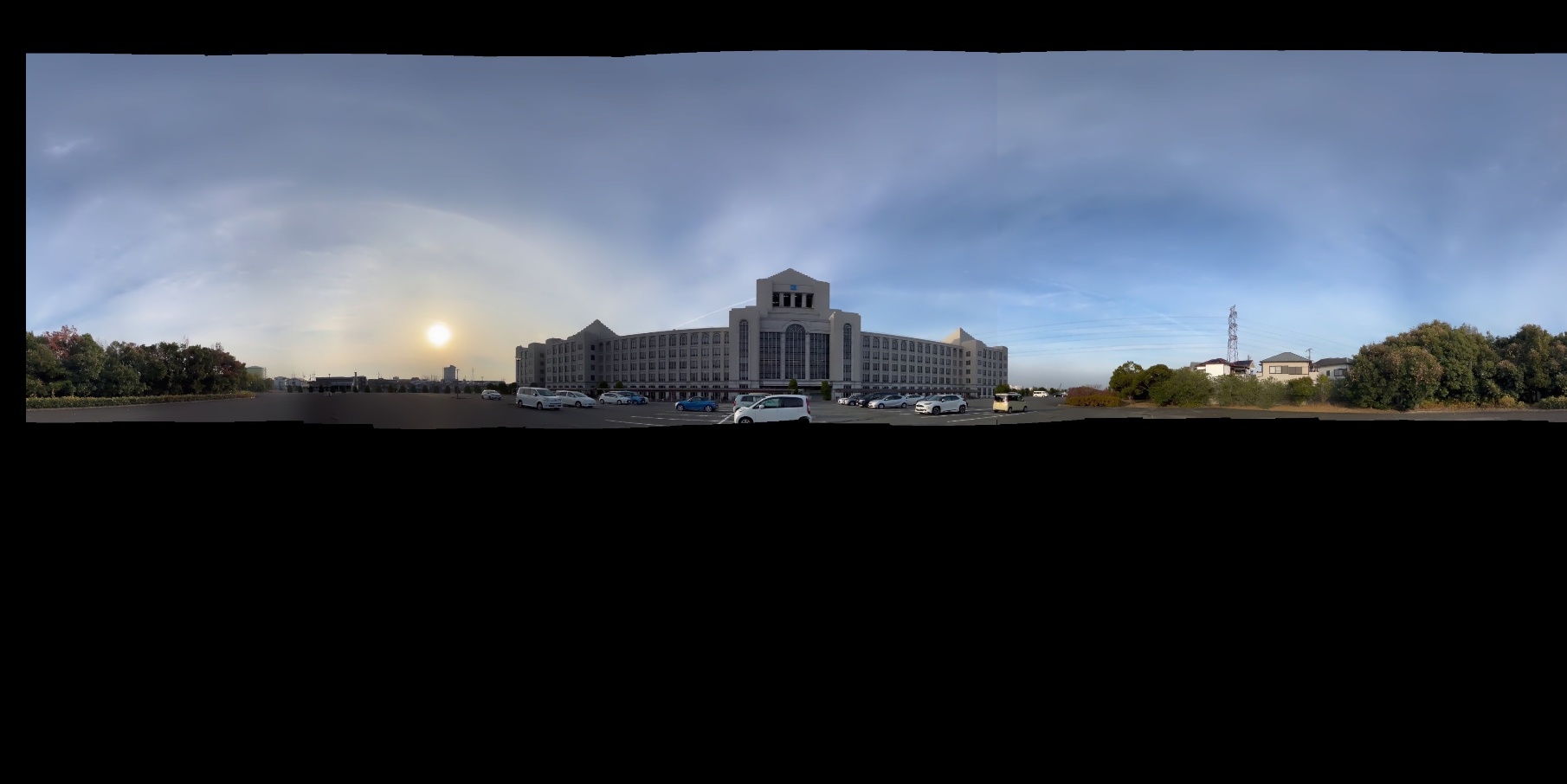}
\subcaption{Transformed image}
\end{minipage}\hspace{1mm}
\begin{minipage}{0.48\textwidth}
\includegraphics[width=\textwidth]{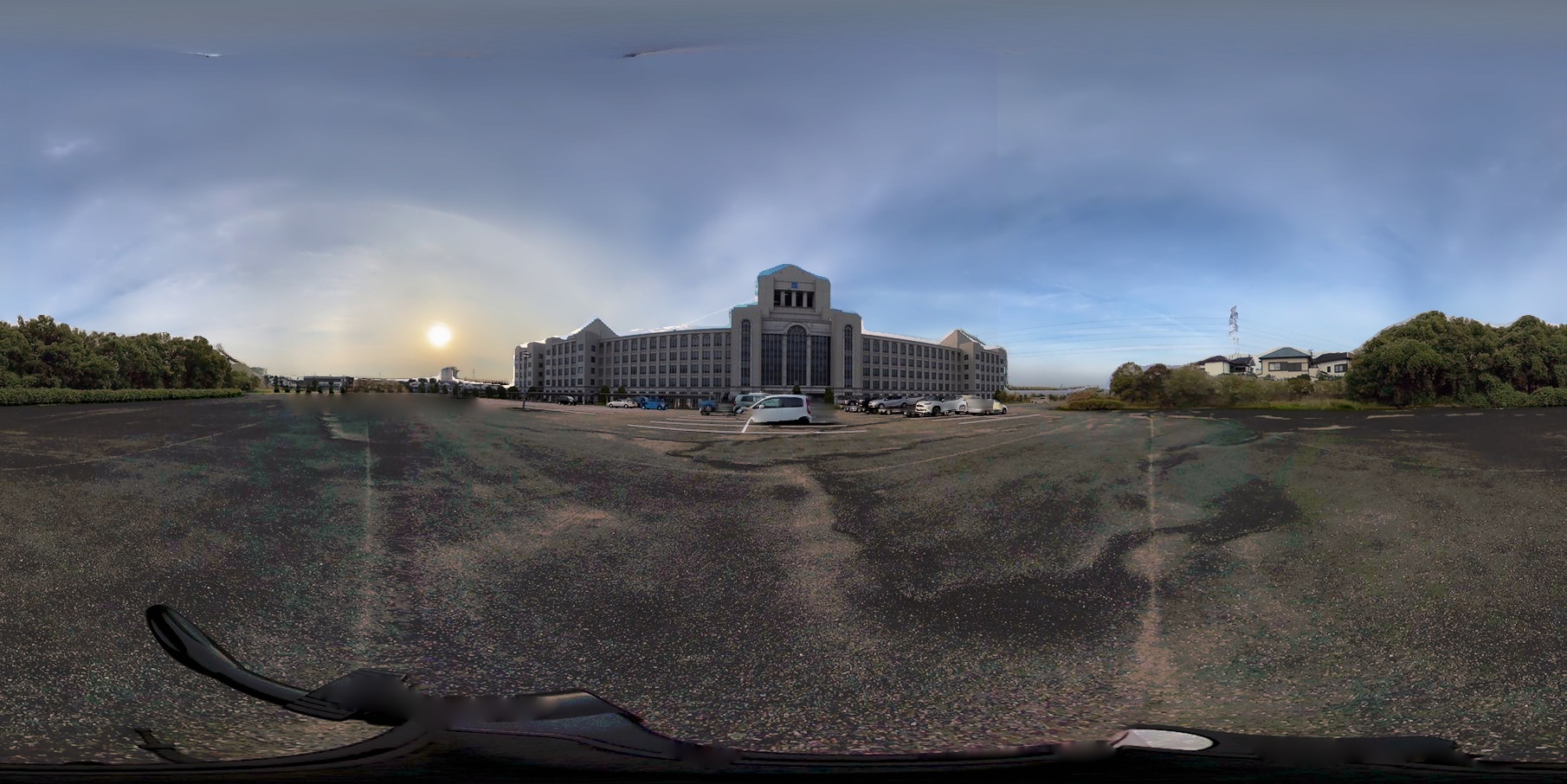}
\subcaption{Result}
\end{minipage}\\
\caption{Experiment 1 on a sunny day for Location 1.}
\label{fig:experiment1}
\end{center}
\end{figure}

\begin{figure}[tb]
\begin{center}
\includegraphics[width=0.36\textwidth]{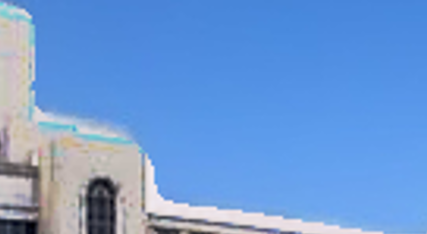}
\caption{Unnatural boundary between sky area and other areas.}
\label{fig:boundary}
\end{center}
\end{figure}

In Experiment 1, we used the pre-captured omnidirectional image as shown in Fig. \ref{fig:pre-captureimages}(a). Figure \ref{fig:experiment1}(a) shows the panoramic image converted from a video captured by the mobile device in a sunny day. The panoramic image was transformed to Fig. \ref{fig:experiment1}(b).
In the figure, the positions of the objects in the scene are almost the same. Using the transformed image, we obtained the intensity and texture-corrected image as shown in Fig. \ref{fig:experiment1}(c). 
From the result, we confirmed that the result is much closer to the current scene than the pre-captured image.
However, as shown in Fig. \ref{fig:boundary}, the boundary between the sky area and the rest of the image remained unnatural. 
This is because of the inaccuracy of the semantic segmentation.

\begin{figure}[tp]
\begin{center}
\begin{minipage}{0.48\textwidth}
\includegraphics[width=\textwidth]{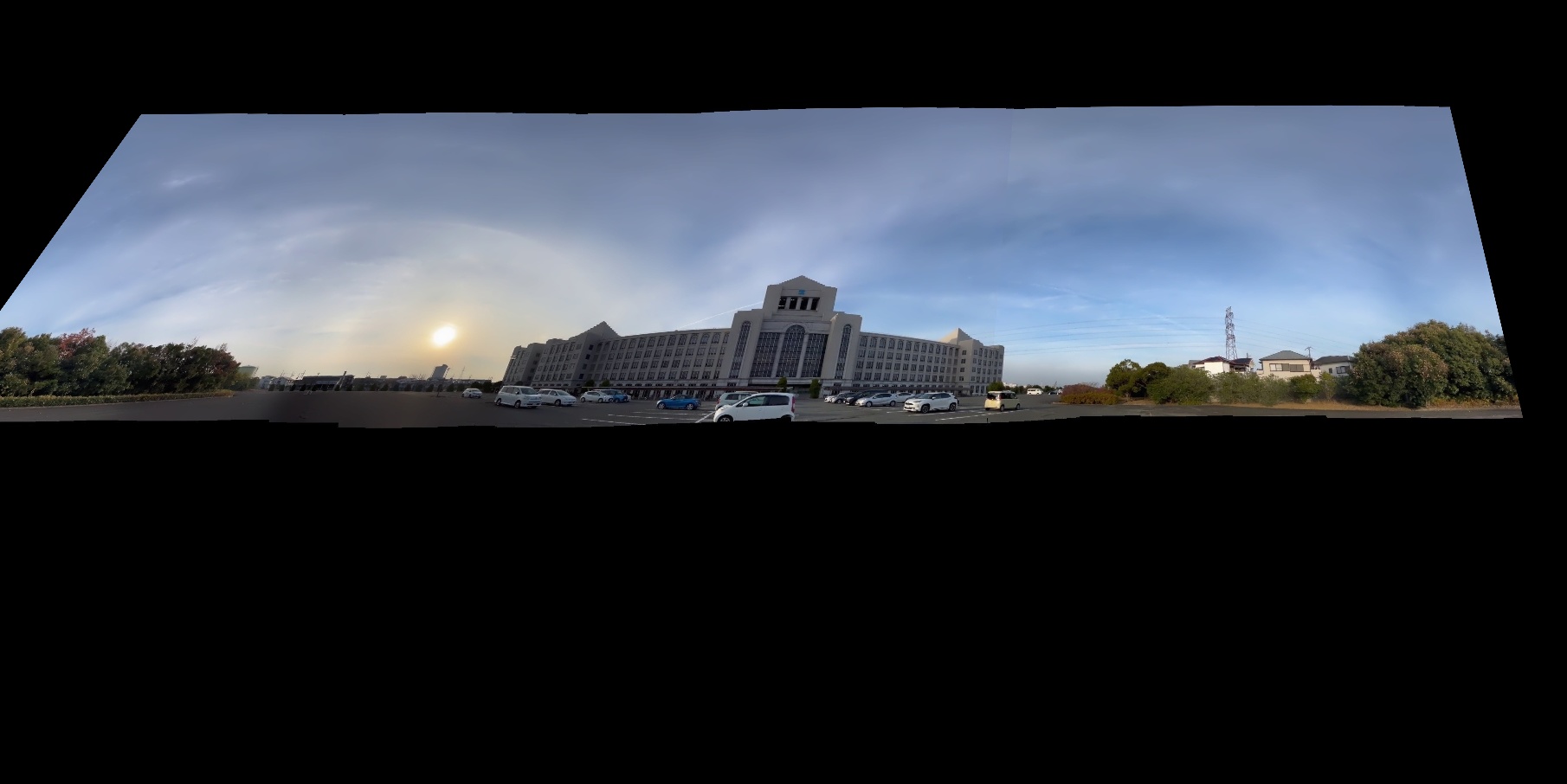}
\subcaption{Result of homography transformation}
\end{minipage} \hspace{1mm}
\begin{minipage}{0.48\textwidth}
\includegraphics[width=\textwidth]{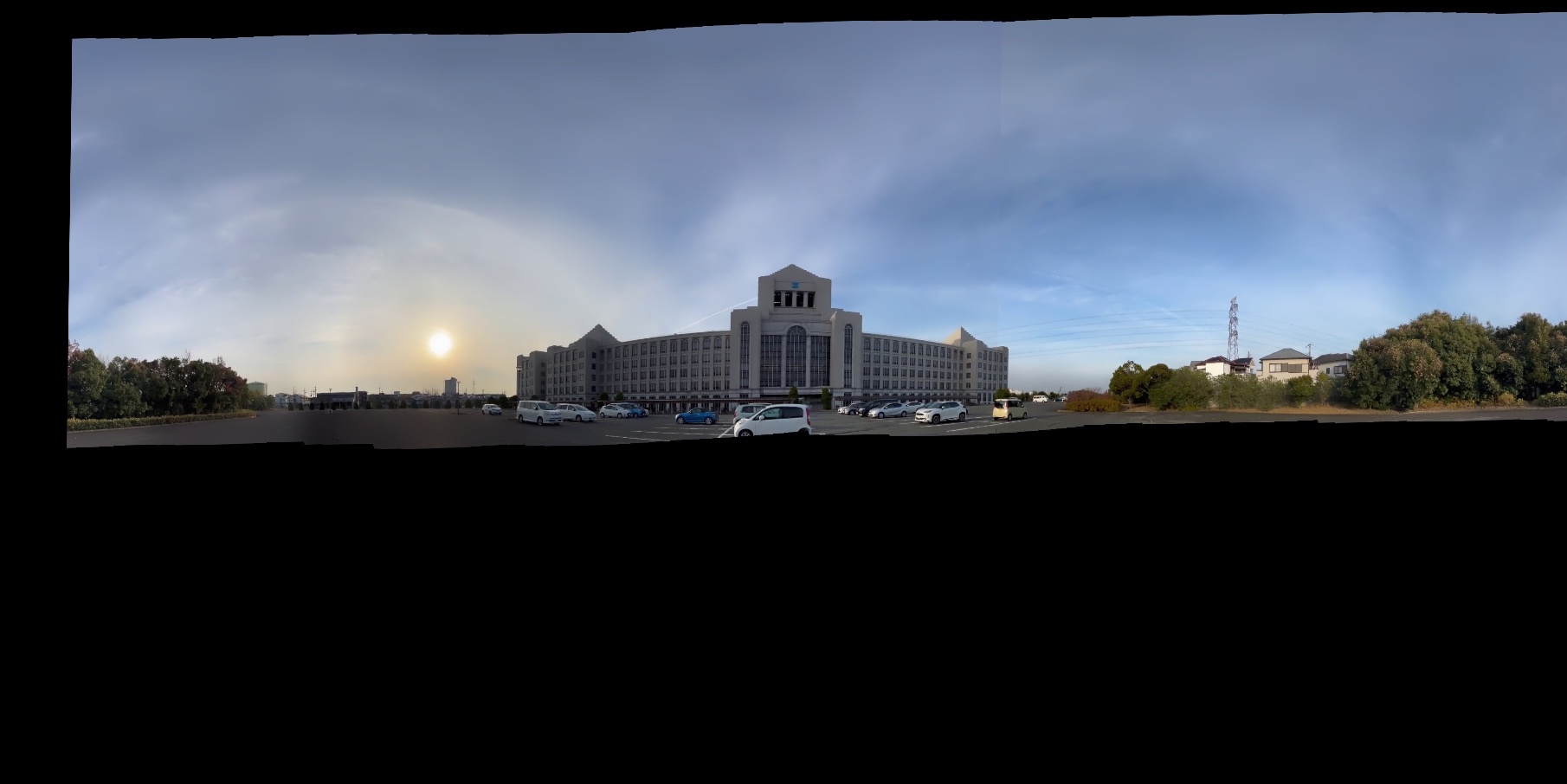}
\subcaption{Result of affine transformation}
\end{minipage}\\
\caption{Examination of other transformation methods.}
\label{fig:transformation}
\end{center}
\end{figure}

In this experiment, we also examined two more transformation methods for aligning panoramic and omnidirectional images. Figure \ref{fig:transformation} shows the results by homography and affine transformations with RANSAC \cite{ransac} from Fig. \ref{fig:experiment1}(a). The images transformed by homography and affine are highly distorted or overall tilted, respectively. From this result, since homography and affine transformations have too many degrees of freedom and distort the image unnecessarily, scaling and translation were found to be the most suitable transformations in this case.

\subsection{Experiments 2}

\begin{figure}[tp]
\begin{center}
\begin{minipage}{0.48\textwidth}
\includegraphics[width=\textwidth]{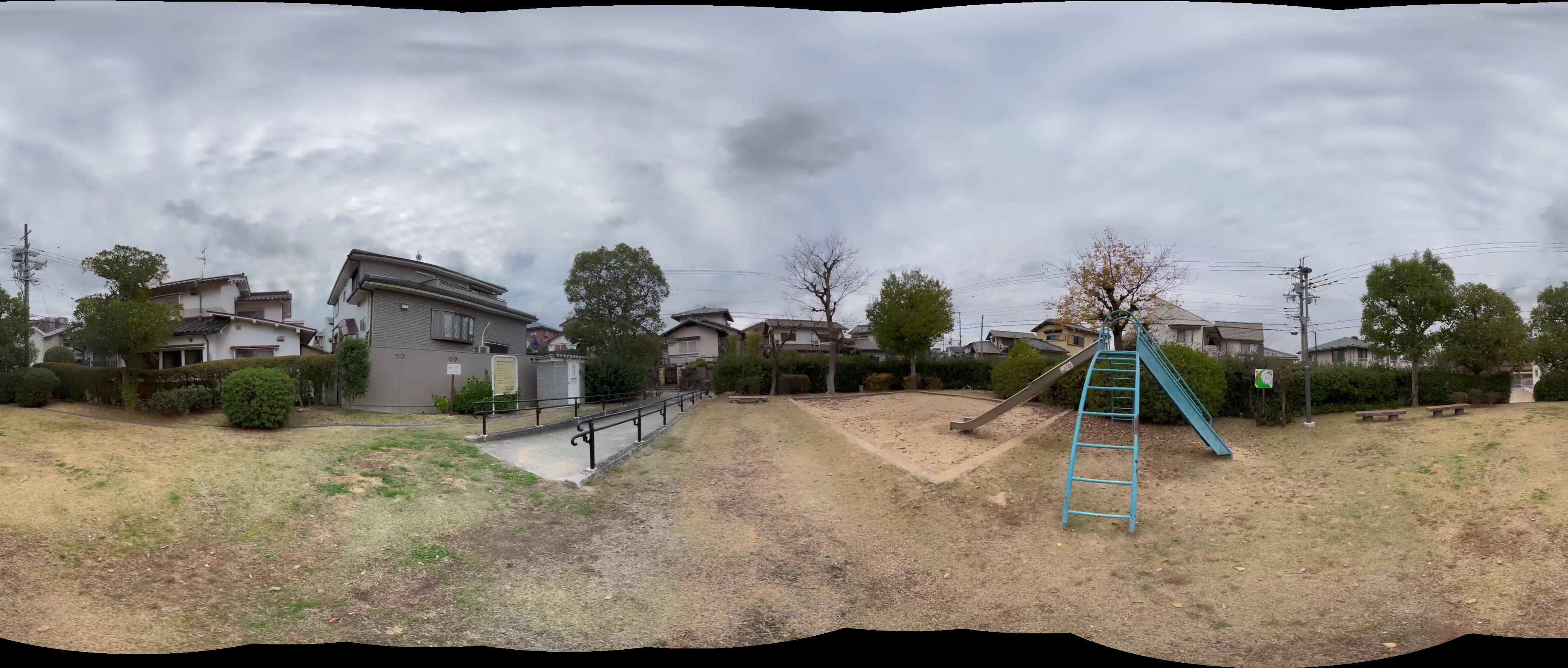}
\subcaption{Panoramic image in Cloudy}
\end{minipage} \\
\begin{minipage}{0.48\textwidth}
\includegraphics[width=\textwidth]{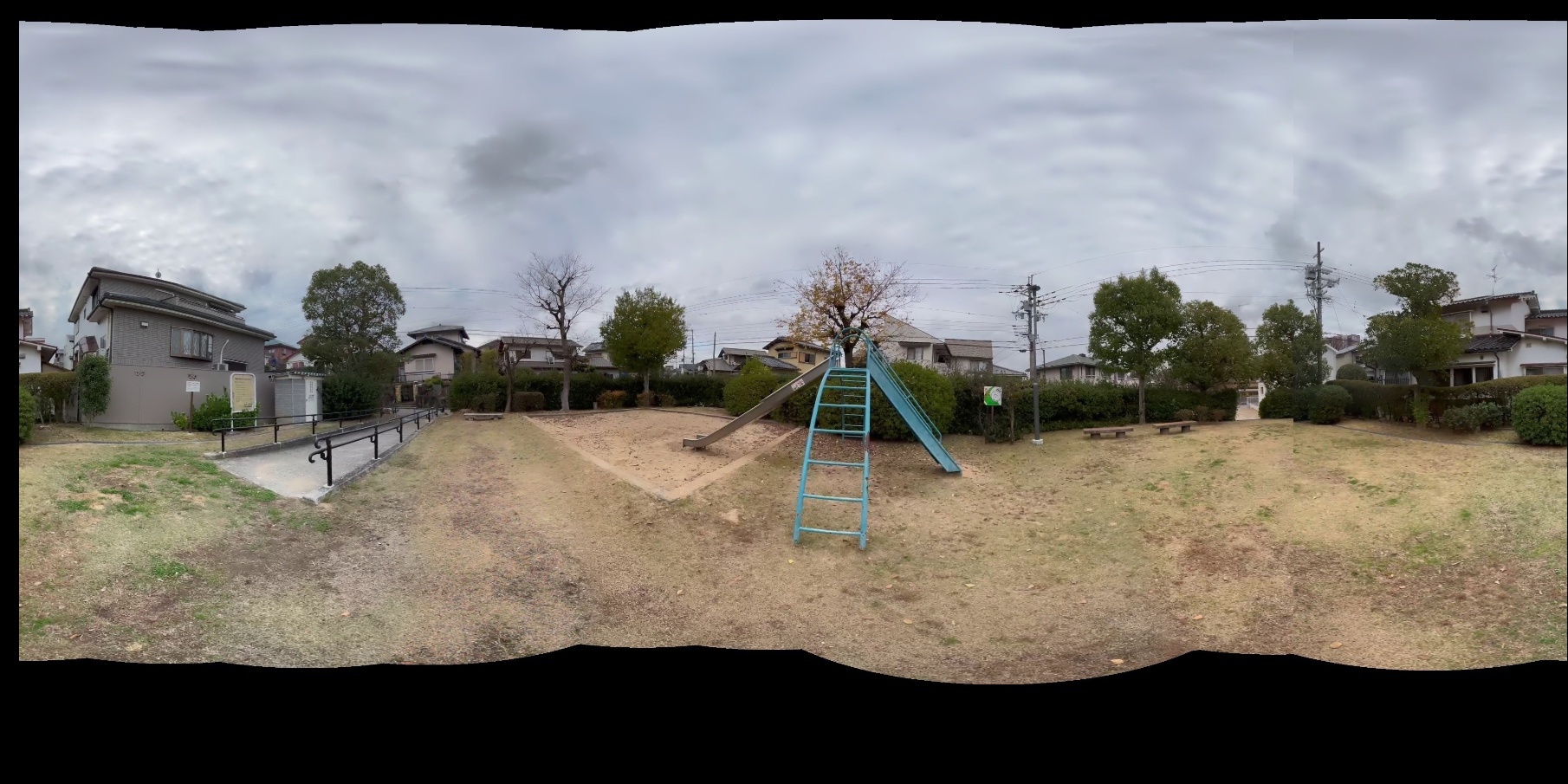}
\subcaption{Transformed image}
\end{minipage}\hspace{1mm}
\begin{minipage}{0.48\textwidth}
\includegraphics[width=\textwidth]{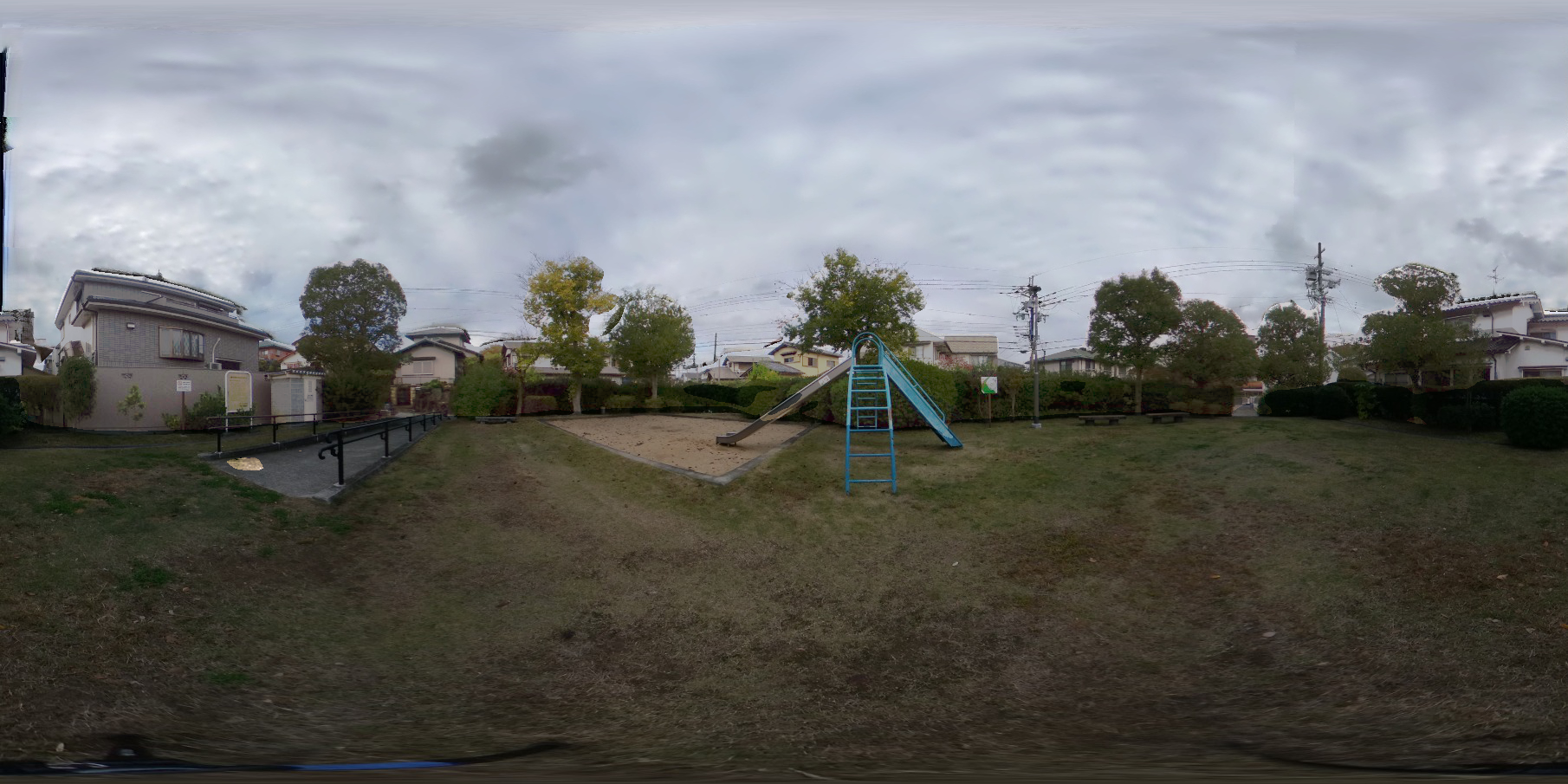}
\subcaption{Result}
\end{minipage}\\
\caption{Experiment 2 on a cloudy day for Location 2.}
\label{fig:experiment2}
\end{center}
\end{figure}

\begin{figure}[tp]
\begin{center}
\begin{minipage}{0.4\textwidth}
\includegraphics[width=\textwidth]{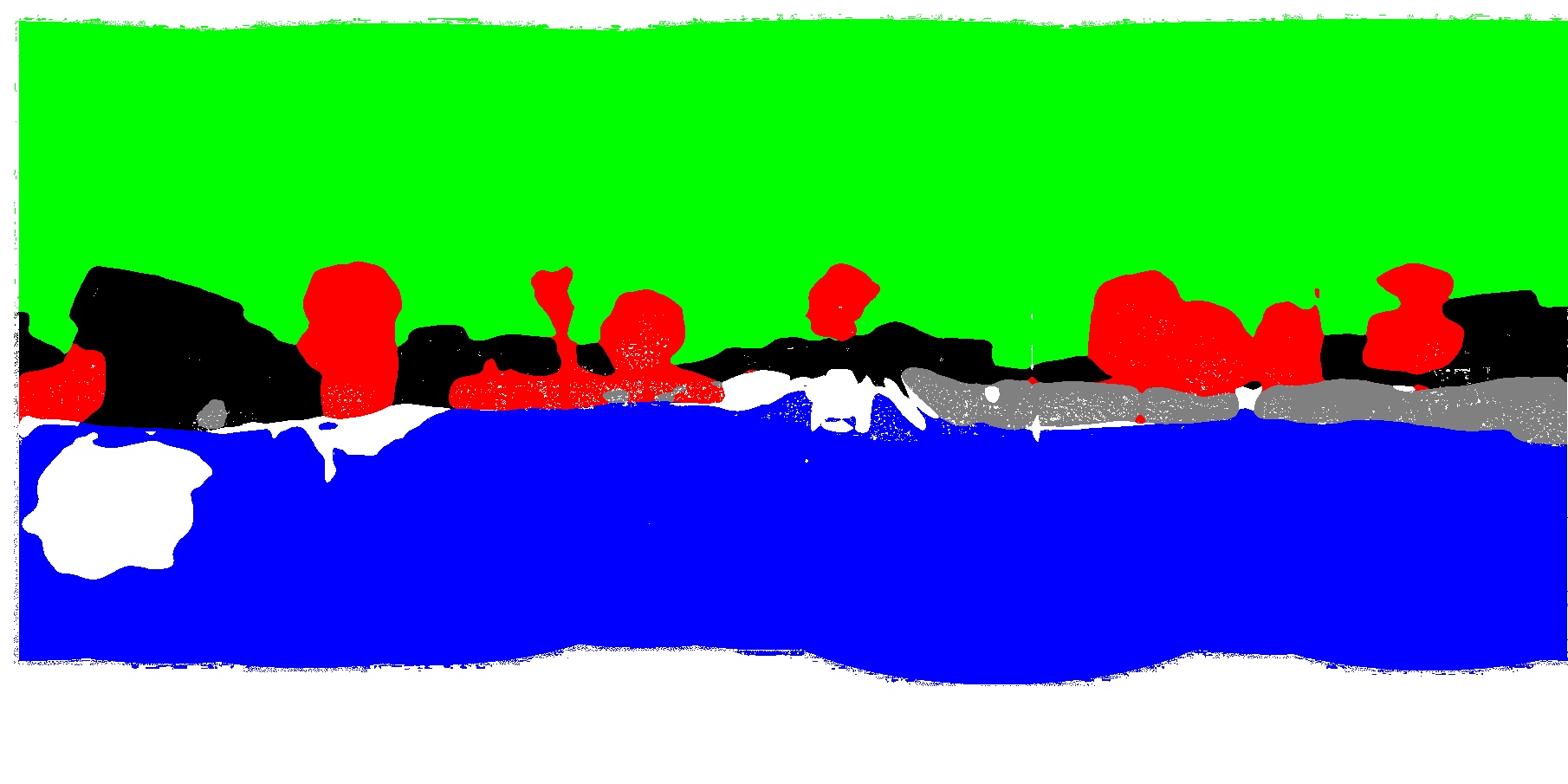}
\subcaption{Labeling of transformed image}
\end{minipage}\hspace{1mm}
\begin{minipage}{0.4\textwidth}
\includegraphics[width=\textwidth]{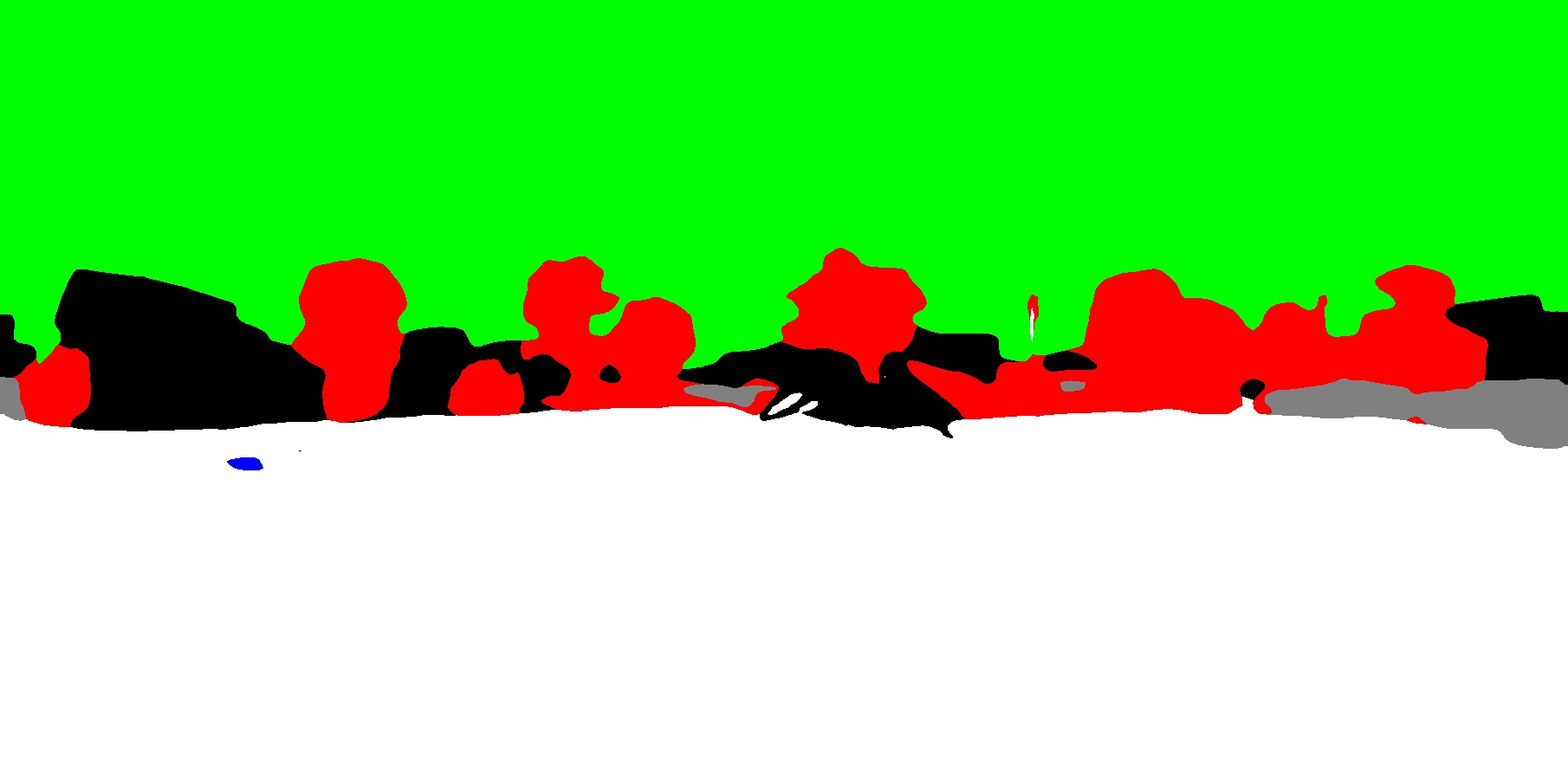}
\subcaption{Labeling of pre-captured image}
\end{minipage} \\
\caption{Labeling images of Experiments 2.}
\label{fig:labelingimages2}
\end{center}
\end{figure}

We conducted Experiment 2 using the pre-captured image at location 2 as shown in Fig. \ref{fig:pre-captureimages}(b) and the panoramic image captured on a cloudy day as shown in Fig. \ref{fig:experiment2}(a). The transformed image shown in Fig. \ref{fig:experiment2}(b) was successfully aligned with Fig. \ref{fig:pre-captureimages}(b).
In the result shown in Fig. \ref{fig:experiment2}(c), the proposed method produced the texture in the sky area closer to the current scene than the pre-captured image.
However, the intensity in the lower half of the image shown in Fig. \ref{fig:experiment2}(c) was hardly corrected.
Figure \ref{fig:labelingimages2} shows the top five categories obtained by semantic segmentation. The categories are: sky (green), tree (red), building (black), earth (blue), plant (gray), and other areas (white). As can be seen in the figure, the ground in the transformed image was classified as earth, while the ground in the pre-capture image was classified as grass, which are different categories though they are the same area. 
In addition, the other areas (white) in the pre-captured image were too large to successfully complete the luminance adjustment of the areas by poisson image editing.
From this fact, we confirmed that if the change in landscape is too large, classification by the semantic partitioning may not be accurate and the proposed method may not work as expected.

\subsection{Experiments 3}
\begin{figure}[tb]
\begin{center}
\begin{minipage}{0.48\textwidth}
\includegraphics[width=\textwidth]{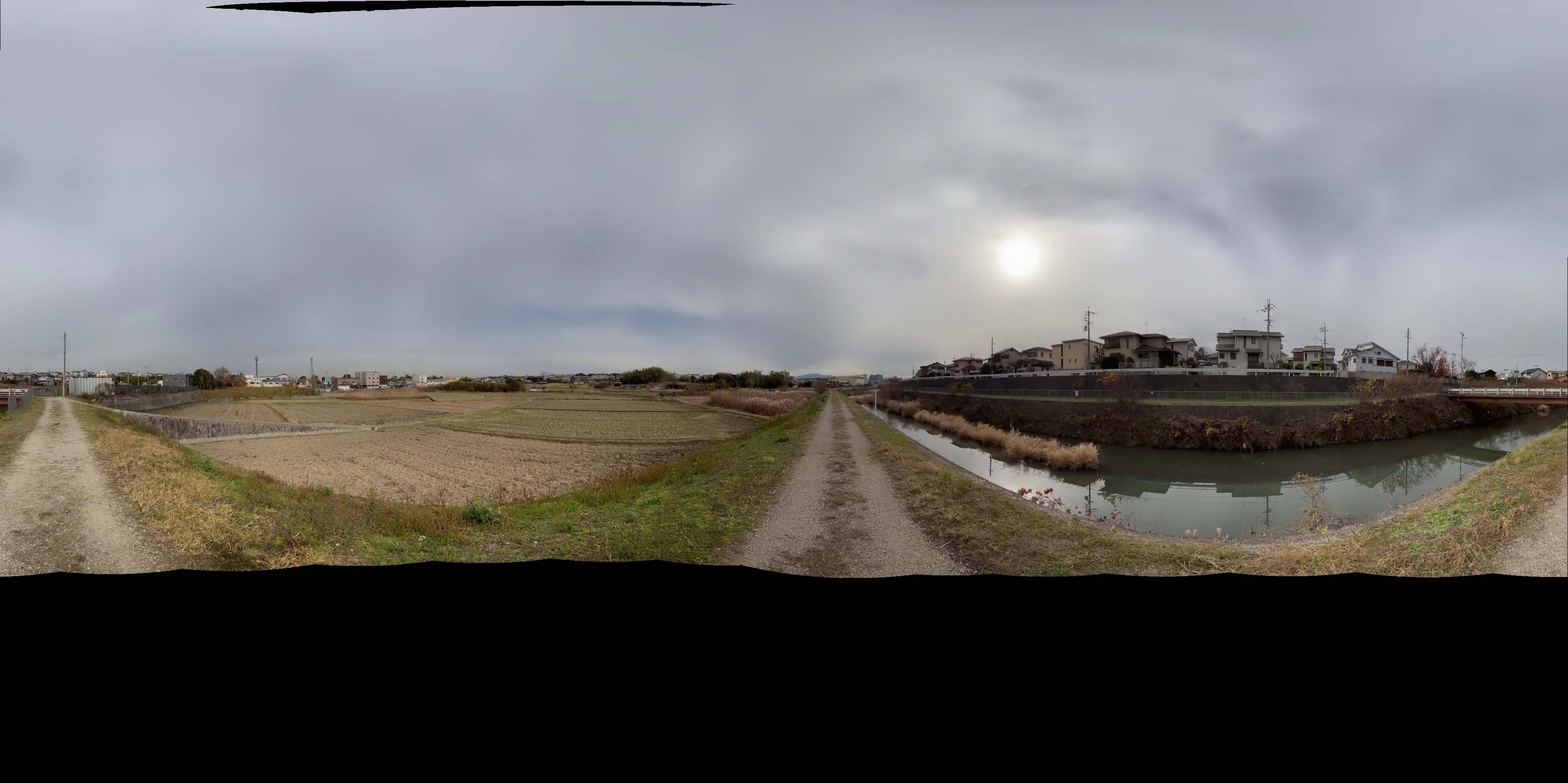}
\subcaption{Panoramic image in Sunny}
\end{minipage} \\
\begin{minipage}{0.48\textwidth}
\includegraphics[width=\textwidth]{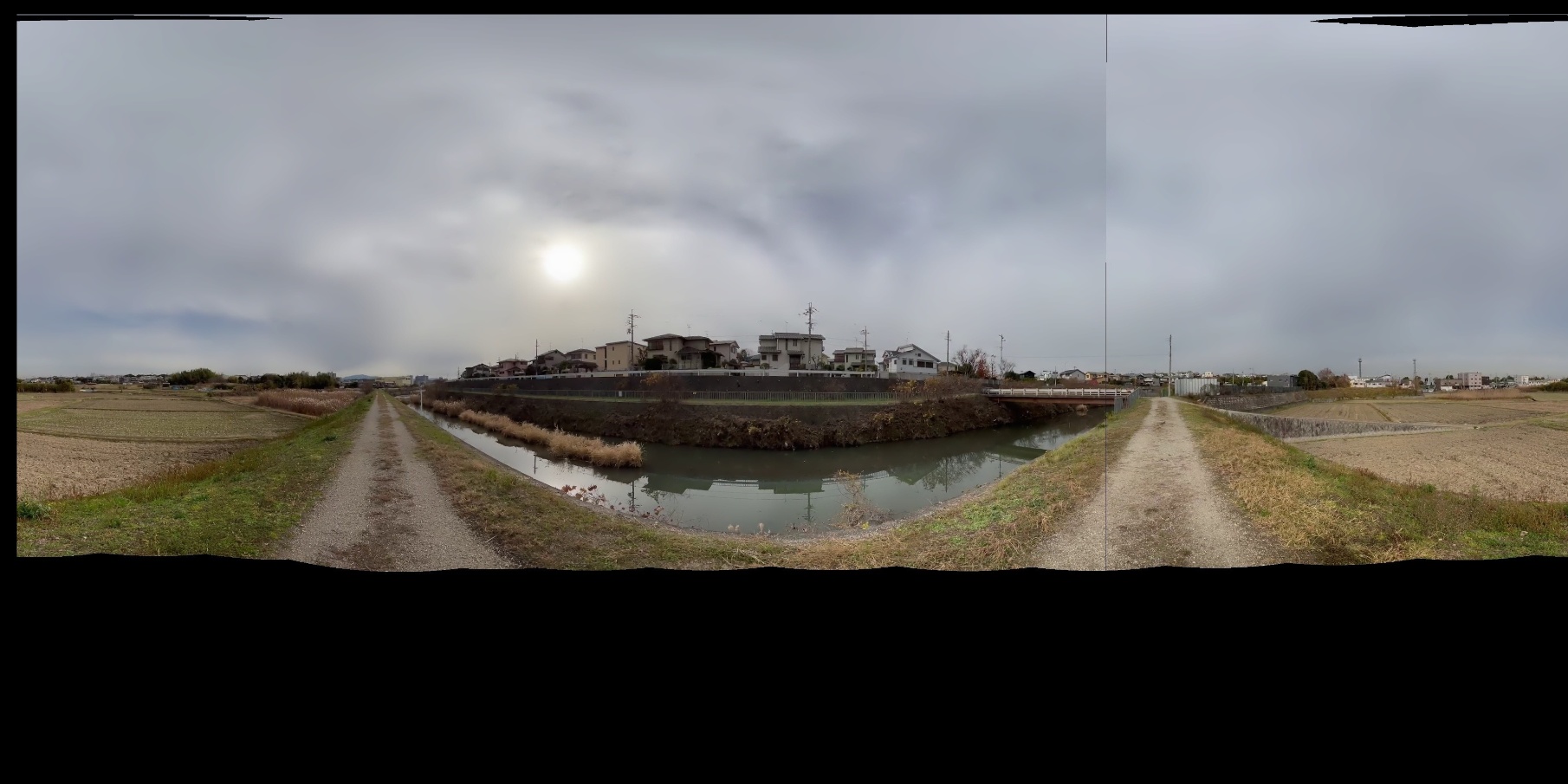}
\subcaption{Transformed image}
\end{minipage}\hspace{1mm}
\begin{minipage}{0.48\textwidth}
\includegraphics[width=\textwidth]{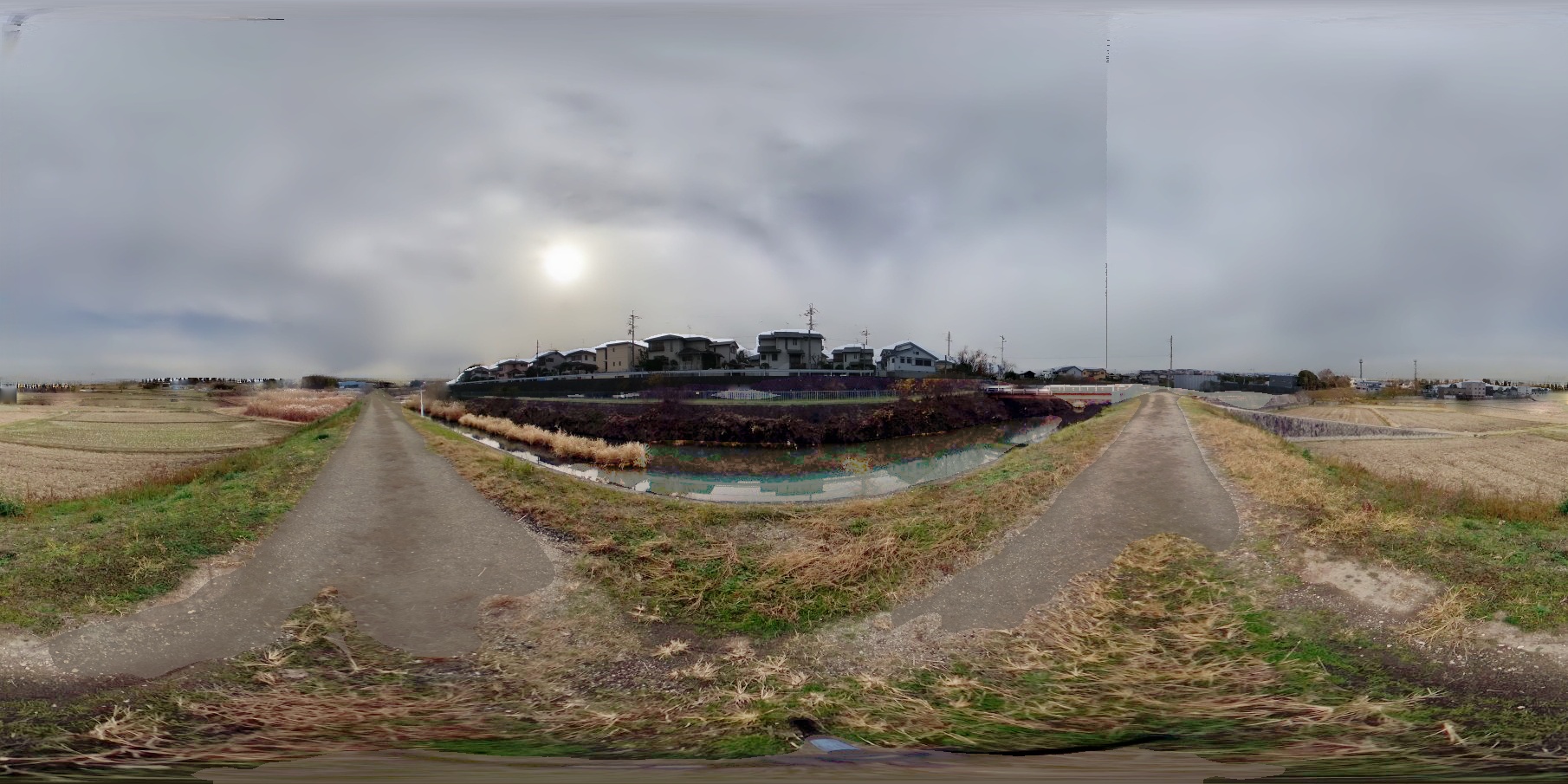}
\subcaption{Sunny evening in Scene 3}
\end{minipage}\\
\caption{Experiment 3 on a sunny morning day for Location 3.}
\label{fig:experiment3}
\end{center}
\end{figure}

\begin{figure}[tb]
\begin{center}
\includegraphics[width=0.48 \textwidth]{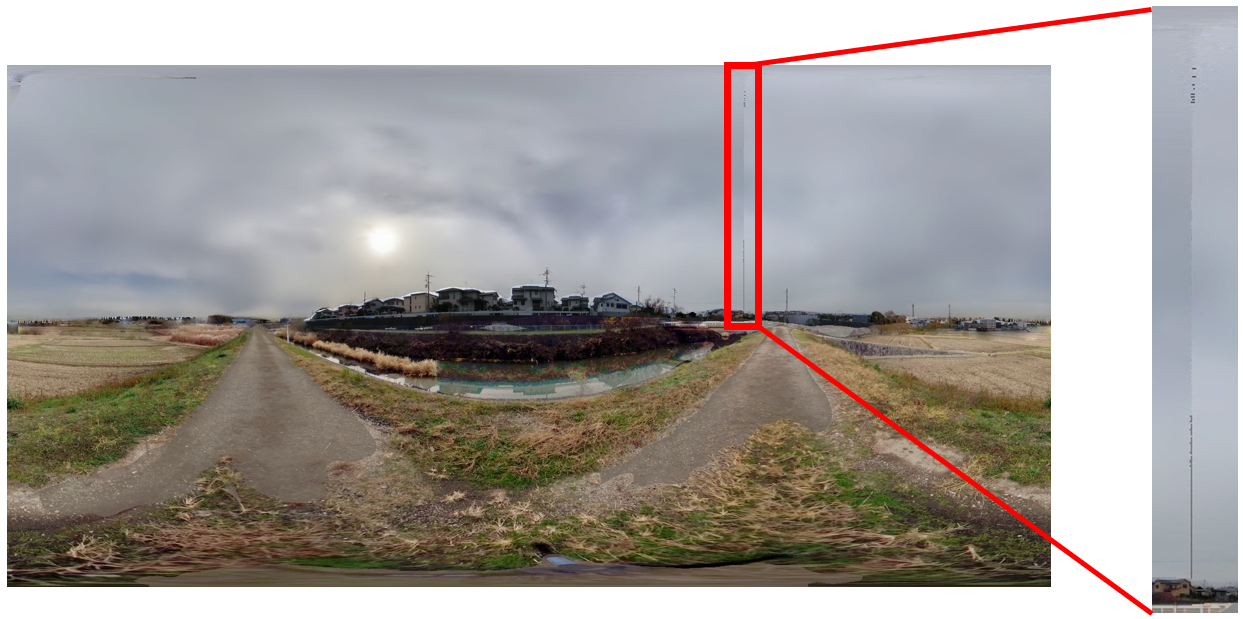}
\caption{Defects in panoramic image creation.}
\label{fig:stitcher}
\end{center}
\end{figure}

We conducted Experiment 3 using the pre-captured image at location 3 as shown in Fig. \ref{fig:pre-captureimages}(c) and the panoramic image captured on a sunny morning as shown in Fig. \ref{fig:experiment3}(a).
The resulting image is more closer to the current landscape, especially in terms of brightness, than the pre-capture image shown in Fig. \ref{fig:experiment3}(c).
However, as shown in Fig. \ref{fig:stitcher}, an unnatural edge occurs in the sky area.
This edge is caused by the fact that when areas at both ends of a panoramic image overlap, the pixel values of one of the areas are simply used in the transformation of the panoramic image. Therefore, it is necessary to improve the method of determining the pixel values in the case of overlap.

\section{Conclusion}
This paper proposed a method to correct the intensity and texture of a pre-captured omnidirectional image using mobile device camera images for Indirect AR.
In the proposed method, a panoramic image was generated from a video taken with a mobile device at the AR experience point, and aligned with the pre-captured omnidirectional image. Using the aligned panoramic image, the intensity and texture of the pre-captured image was corrected based on histogram matching, poisson image editing, texture composition, and inpainting.
In our experiments, we succeeded in correcting the intensity and texture in some cases, but we also confirmed that the accuracy and the classification categories of semantic segmentation have a significant impact on the results.
Future work includes the introduction of more accurate semantic segmentation and the investigation of the IAR experience using the corrected images.

\subsubsection*{Acknowledgements}
This work was partially supported by JSPS KAKENHI Grant Number JP21H03483.

%
%

%
%
%
%
\bibliographystyle{splncs04}
\bibliography{egbib}

\end{document}